\pdfoutput=1

\documentclass[11pt]{article}

\usepackage[]{acl}

\usepackage{times}
\usepackage{latexsym}

\usepackage[T1]{fontenc}

\usepackage[utf8]{inputenc}

\usepackage{microtype}

\usepackage{soul} 
\newcommand{\drop}[1]{\colorbox{red!10}{#1}}
\newcommand{\best}[1]{\colorbox{black!10}{#1}}

\usepackage{tipa}
\usepackage{dblfloatfix}
\usepackage{amsmath}
\usepackage{graphicx}
\usepackage{array,booktabs,ragged2e}
\usepackage{multirow}
\newcolumntype{R}[1]{>{\RaggedLeft\arraybackslash}p{#1}}
\newcolumntype{L}[1]{>{\RaggedRight\arraybackslash}p{#1}}
\newcolumntype{M}[1]{>{\centering\arraybackslash}m{#1}}
\usepackage{xcolor}

\newcommand{\data}[1]{VALUE#1}
\newcommand{\org}[1]{DataWorks#1}
\newcommand{\workers}[1]{DataWorkers#1}
\newcommand{\numAAVEFeatures}{11}

\title{{VALUE}: {U}nderstanding Dialect Disparity in {NLU}}

\author{Caleb Ziems \hspace{1.5em}
        Jiaao Chen \hspace{1.5em}
        Camille Harris\\
        \textbf{Jessica Anderson} \hspace{1.5em}
        \textbf{Diyi Yang} \\
        Georgia Institute of Technology\\
        \texttt{\{\href{mailto://cziems3@gatech.edu}{cziems}, \href{mailto://jchen896@gatech.edu}{jchen896}, \href{mailto://charris320@gatech.edu}{charris320}\}@gatech.edu}\\
        \texttt{\href{mailto://diyiy@cs.stanford.edu}{diyiy@cs.stanford.edu}}
}

\date{}

\begin{document}
\maketitle
\begin{abstract}
English Natural Language Understanding (NLU) systems have achieved great performances and even outperformed humans on benchmarks like GLUE and SuperGLUE. However, these benchmarks contain only textbook Standard American English (SAE). Other dialects have been largely overlooked in the NLP community. This leads to biased and inequitable NLU systems that serve only a sub-population of speakers. To understand disparities in current models and to facilitate more dialect-competent NLU systems, we introduce the VernAcular Language Understanding Evaluation (\data) benchmark, a challenging variant of GLUE that we created with a set of lexical and morphosyntactic transformation rules. In this initial release (V.1), we construct rules for \numAAVEFeatures{} features of African American Vernacular English (AAVE), and we recruit fluent AAVE speakers to validate each feature transformation via linguistic acceptability judgments in a participatory design manner. Experiments show that these new dialectal features can lead to a drop in model performance. To run the transformation code and download both synthetic and gold-standard dialectal GLUE benchmarks, see \url{https://github.com/SALT-NLP/value}
\end{abstract}

\section{Introduction}
Most of today's research in NLP mainly focuses on 10 to 20 high-resource languages with a special focus on English, though there are thousands of languages and dialects with billions of speakers in the world. 
NLU systems that are trained on polished or ``textbook'' Standard American English (SAE) are not as robust to linguistic variation \cite{belinkov2017synthetic,ebrahimi2018hotflip}.
While some recent works have challenged leading systems with adversarial examples like typos \cite{jones2020robust}, syntactic rearrangements \cite{iyyer2018adversarial}, and sentence/word substitutions \cite{alzantot2018generating,jia2017adversarial,ribeiro2018semantically}, fewer have considered the effects of dialectal differences on performance. When language technologies are not built to handle dialectal differences, the benefits of these technologies may not be equitably distributed among different demographic groups \cite{hovy2016social}. Specifically, models tested on African American Vernacular English (AAVE) have been found to struggle with language identification \cite{jurgens2017incorporating}, sentiment analysis \cite{kiritchenko2018examining}, POS tagging \cite{jorgensen2016learning} and dependency parsing \cite{blodgett2018twitter},
and led to severe racial disparities in the resulting language technologies 
such as the automated speech recognition used by virtual assistants \cite{koenecke2020racial}  the hate speech detection used by online media platforms \cite{rios2020fuzze,halevy2021mitigating}.

However, no prior work has systematically investigated these dialect-specific shortcomings across a broad set of NLU tasks, 
and the effectiveness of low-resource NLP methods for dialectal Natural Language Understanding (NLU) remains largely unexplored. The first barrier to progress is that a standard benchmark for dialectal NLU has not yet been constructed. The second is that no systematic error analyses have yet revealed causal insights about the specific challenges that models face with domain adaptation to different language varieties. 

To both understand dialect disparity and facilitate ongoing work on dialect-competent NLU, we introduce a new dialect-specific challenge dataset -- the \textbf{V}ern\textbf{A}cular \textbf{L}anguage \textbf{U}nderstanding \textbf{E}valuation benchmark (\textbf{\data}).  
We specifically focus on African American Vernagular English (AAVE), a dialect spoken by nearly 33 million people, and approximately 80\% of African Americans in the United States \cite{lippi1997we}. To facilitate  direct comparison with prior work, we build \data{} by directly transforming GLUE \cite{wang2018glue} into synthetic AAVE.

Our AAVE transformation pipeline comes with two key advantages: it is flexible enough to facilitate an interpretable perturbation error analysis, and the transformation rules are meaning-preserving, which ensures the validity of the transformed NLU tasks. Our pipeline includes a set of linguistically-attested rules for syntax (sentence structure; e.g. negation rules), morphology (word structure; e.g., suffixes), orthography (writing and spelling conventions), and the lexicon (the list of available words and phrases). Because our system is rule-based, we can isolate and systematically test which features most significantly challenge models. While it is also possible to generate pseudo-dialects via end-to-end style transfer \cite{krishna2020reformulating}, these systems often fail to disentangle style from content, and thus also fail to preserve meaning \cite{lample2018multiple}. We confirm these shortcomings in this work, and  affirm the validity of our own meaning-preserving transformation rules via the acceptability judgments of fluent AAVE speakers in a participatory design manner. 
To sum up, our work contributes the following: 
\begin{enumerate}\setlength\itemsep{0em}
    \item \textbf{Dialect Transformations:} A set of \numAAVEFeatures{} new linguistic rules for reliably transforming Standard American English (SAE) into African American Vernacular English (AAVE).
    \item \textbf{\data{}}: An AAVE benchmark dataset with seven NLU tasks.
    \item \textbf{Synthetic + Gold Standard Data:} Robust validation of synthetic transformations as well as gold standard dialectal data from native AAVE speakers via an iterative participatory design process.
    \item \textbf{Benchmark Evaluation:} Experiments with RoBERTA baselines plus fine-tuning methods to improve model robustness on dialectal variants.
    \item \textbf{Dialect-Specific Analysis:} Perturbation analysis that reveals the task-specific challenges of AAVE-specific grammatical features.
\end{enumerate}

\section{Related Work}
\paragraph{Computational Sociolinguistics of Dialect} 
Prior work on developing NLU models has often used dominant English varieties, Standard American English (SAE), owing to the availability of text datasets for training and testing \cite{blodgett2016demographic}.  
Models can marginalize certain groups when trained on datasets that lack linguistic diversity or contain biases against minority language speakers 
\cite{DBLP:journals/corr/BlodgettO17}.
Despite these shortcomings, there still has been relatively little attention paid to dialects in the language technologies research communities.
Prior studies have mainly focused on distinguishing between English language varieties \cite{demszky-etal-2021-learning,zampieri2014report}.

Failure to account for dialects like AAVE can lead to performance degradation of the NLU tools such as Automatic Speech Recognition (ASR) \cite{dorn2019dialect}, Language Identification (LID) and dependency parsing tools \cite{blodgett2016demographic}. \citet{hwang-etal-2020-towards}
also demonstrated the inadequacy of WordNet and ConceptNet in reflecting AAVE and other varieties.
Thus there have been several works highlighting the need for AAVE-inclusivity in NLU \cite{groenwold2020investigating}. Despite its large community of speakers, AAVE is under-represented in current technologies.

\paragraph{Model Robustness and Challenge Datasets} Language technologies are not inherently robust to linguistic variation. The performance of neural models is expected to degrade due to sparsity in the presence of non-canonical text \cite{zalmout2018noise,belinkov2017synthetic,ebrahimi2018hotflip}, as shown empirically for random character, word, and sentence-level permutations \cite{jones2020robust,alzantot2018generating,jia2017adversarial,ribeiro2018semantically,iyyer2018adversarial}. This has motivated growing interest in challenging datasets based on adversarial perturbations \cite{nie-etal-2020-adversarial,tan2020s}, spurious patterns or correlations \cite{zhang-etal-2019-paws,mccoy-etal-2019-right}, and counterfactual examples \cite{gardner-etal-2020-evaluating,Kaushik2020Learning}.  
However, the same attention has not been shown to dialects, which vary \textit{systematically} in their syntax, morphology, phonology, orthography, and lexicon \cite{jurgens2017incorporating}. To this end, we introduce the evaluation set by adapting from the in-distribution examples (SAE) to out-of-distribution examples (AAVE) on GLUE benchmarks. Our goal is to develop robust models that have a good performance on test sets in different linguistic variations.

\section{Constructing \data}
\label{sec:data}

We constructed \data{} from the widely-used GLUE benchmark \cite{wang2018glue}, which contains NLU tasks such as natural language inference (e.g., MNLI; \citeauthor{bowman2015large}), question answering (QNLI; \citeauthor{rajpurkar2016squad}), and linguistic acceptability (CoLA; \citeauthor{warstadt2018neural}). For each of the main tasks, we translated the Standard American English (SAE) into a synthetic form of AAVE --- a form containing many of AAVE's distinguishing features with extremely high concentration. We implemented these transformations using a set of lexical and morphosyntactic rules derived from a broad survey of the linguistics literature \cite{collins2008aae,green2002african,labov1972language,labov1998co,sidnell2002african,stewart2014now,thompson2016morpho,wolfram2015american}. These features were specifically chosen for their high empirical attestation across regional and generational variants of AAVE.

\subsection{Morphosyntactic Translation} 
\label{subsec:morphosyntactic_translation}

This work represents the first attempt to systematically catalogue and operationalize a set of computational rules for inserting AAVE-specific language structures into text. We distill field linguists's observations into procedural code, which operates on specific grammatical conditions from the SAE source. Each grammatical condition is specified by the part of speech tags and syntactic dependency relationships present in the text. Appendix~\ref{appdx:morphosyntax} lists all implementation details for each transformation rule, and we will now enumerate them briefly.

\paragraph{Auxiliaries.} AAVE allows copula deletion and other auxiliary dropping \cite{stewart2014now,green2002african,labov1972language,wolfram2015american}. This means the SAE sentence ``\textit{We are better than before}'' could be rendered in AAVE without the copula as ``\textit{We better than before}.'' We look for the present tense \textit{is} and \textit{are} as well as any tokens with \texttt{AUX} part of speech tag to drop (under special conditions listed in more detail in Appendix~\ref{appdx:morphosyntax}).

\paragraph{Completive \textit{done} and remote time \textit{been}.} The phrase ``\textit{I had written it.}'' can be rendered in AAVE as ``\textit{I done wrote it}'' using the completive verbal marker \textit{done}. The phrase ``\textit{He ate a long time ago}'' can be rendered as ``\textit{He been ate}'' using the remote time \textit{been} \cite{green2002african}.

\paragraph{Constructions involving the word \textit{ass}.} These constructions may be misclassified as obscenity, but they serve a distinct and consistent role in AAVE grammar \cite{spears1998african}. One common form is called the \textit{ass} camouflage construction \cite{collins2008aae}, and it can be seen in the phrase ``\textit{I divorced his ass}.'' Here, the word behaves as a metonymic pseudo-pronoun \cite{spears1998african}. Similarly, it can appear reflexively, as in ``\textit{Get yo'ass inside}.'' \textit{Ass} constructions can also serve as discourse-level expressive markers or intensifiers, as in the compound ``\textit{We was at some random-ass bar}.''

\paragraph{Existential \textit{dey}/\textit{it}.} AAVE speakers can indicate something exists by using what is known as an \textit{it} or \textit{dey} existential construction \cite{green2002african}. The existential construction in ``\textit{It's some milk in the fridge}'' is used to indicate ``\textit{There is some milk in the fridge}.'' We identify existential dependencies for this transformation.

\paragraph{Future \textit{gonna} and immediate future \textit{finna}.} AAVE speakers can mark future tense with \textit{gon} or \textit{gonna}, as in ``\textit{You gon understand}'' \cite{green2002african,sidnell2002african}. In the first person, this becomes \textit{I'ma}. In the immediate future, speakers can use \textit{finna} (or variants \textit{fixina, fixna and fitna}), as in ``\textit{I'm finna leave.}'' 

\paragraph{Have / got.} In the casual speech of AAVE and other dialects, both the modal and the verb form of \textit{have} can be replaced by \textit{got} \cite{trotta2011game}. \textit{Have to} can become \textit{got to} or \textit{gotta}, and similar for the verb of possession. We simply convert the present-tense \textit{have} and \textit{has} to \textit{got} and ensure that the verb has an object.

\paragraph{Inflection.} In AAVE, speakers do not necessarily inflect simple present or past tense verbs differently for number or person \cite{green2002african}. This means the SAE sentence ``\textit{She studies linguistics}'' could be rendered in AAVE as ``\textit{She \textbf{study} linguistics}.'' We use the \texttt{pyinflect} library to convert all present and simple past verbs into the first person.

\paragraph{Negative concord.} This widely-known feature of AAVE (and numerous other dialects) involves two negative morphemes to convey a single negation.  \cite{martin1998sentence}. For example, the SAE sentence ``\textit{He doesn't have a camera}'' could look more like ``\textit{He don't have no camera}'' in AAVE. This transformation rule is sensitive to the verb-object dependency structure, and requires that the object is an indefinite noun \cite{green2002african}.

\paragraph{Negative inversion.} This feature is superficially similar to negative concord. Both an auxiliary and an indefinite noun phrase are negated at the beginning of a sentence or clause \cite{green2002african,martin1998sentence}. For example, the SAE assertion that ``\textit{no suffering lasts forever}'' could be rendered in AAVE as ``\textit{don't no suffering last forever.}''

\paragraph{Null genitives.} AAVE allows a null genitive marking \cite{stewart2014now,wolfram2015american}, like the removal of the possessive \textit{'s} in ``\textit{Rolanda bed}'' \cite{green2002african}. We simply drop any possessive endings (\texttt{POS}) from the text.

\paragraph{Relative clause structures.} There is a grammatical option to drop the Wh-pronoun when it is serving as the complementizer to a relative clause, as in ``\textit{It's a whole lot of people \textbf{\O} don't wanna go to hell}'' \cite{green2002african}. In our transformation, we simply drop all lemmas \textit{who} and \textit{that} where the head is a relative clause modifier.

\subsection{Lexical and Orthographic Translation} 
Some of the most recognizable differences between SAE and AAVE are found in the lexicon and orthographic conventions. Because we are not aware of any comprehensive AAVE lexicons, we automatically learn our own SAE to AAVE dictionary from public data, and we will provide this resource in our public repository. This dictionary serves as a mapping between plausible synonyms (e.g., \textit{mash}/\textit{press}; \textit{homie}/\textit{friend}; \textit{paper}/\textit{money}) and orthographic variants (e.g., \textit{da}/\textit{the}; \textit{wit}/\textit{with}; \textit{sista}/\textit{sister}) 

In a method inspired by \citet{shoemark2018inducing}, we trained a skip-gram word embedding model\footnote{We used \texttt{gensim} word2vec with dimension $d=200$} \cite{mikolov2013distributed} on the public \texttt{TwitterAAE} dataset of \citet{blodgett2016demographic}. This dataset contained attested code-switching behavior, which allowed us to extract a \textit{linguistic code} axis $c$ in the embedding space, defined by the average
\begin{equation*}
    \boldsymbol{c} = \sum_{(\boldsymbol{x}_i, \boldsymbol{y}_i) \in S} \frac{\boldsymbol{x}_i - \boldsymbol{y}_i}{|S|}
\end{equation*}
where $S$ was our seed list of known priors from \citet{shoemark2018inducing}, given in Appendix~\ref{appdx:lexicon}. 

Next, we ranked the candidate word pairs $\boldsymbol{w}_i, \boldsymbol{w}_j$ by $\cos(\boldsymbol{c}, \boldsymbol{w}_i-\boldsymbol{w}_j)$ following \citet{bolukbasi2016man}. In this ranking, we consider only the pairs whose cosine similarity passed a threshold $\delta$, where $\delta$ was defined by the bottom quartile of the cosine similarities in our seed set $S$. After automatic filtering, we were left with 2,460 pairs. We hand-filtered this list to remove any semantically dissimilar words, like \textit{fishin/kayakin} or \textit{mom/gramps}. This left us with 1,988 pairs. 

Note that these pairs are not one-to-one, but a one-to-many dictionary mapping from SAE to AAVE variants. We provide a sample of this mapping in Table~\ref{tab:lexical_mapping}. In the final step of the translation, we chose uniformly at random between the AAVE variants to make our substitution. We simply scanned the GLUE dataset and swapped any known tokens from SAE to AAVE.

\begin{table}[]
    \centering
    \begin{tabular}{ll}
    \toprule
    \textbf{SAE} & \textbf{AAVE} \\ \midrule
    arguing & \textit{beefing, beefin, arguin} \\ \hline
    anymore & \textit{nomore, nomo} \\ \hline
    brother & \textit{homeboy} \\ \hline
    classy & \textit{fly} \\ \hline
    dude & \textit{n*ggah, manee, n*gga} \\ \hline
    huge & \textit{bigass} \\ \hline
    probably & \textit{prob, prolly, def, probly, deff} \\ \hline
    rad & \textit{dope} \\ \hline
    remember & \textit{rememba} \\ \hline
    screaming & \textit{screamin, yellin, hollering} \\ \hline
    sister & \textit{sista, sis} \\ \hline
    these & \textit{dese, dem} \\ \hline
    with & \textit{wit} \\
    \bottomrule
    \end{tabular}
    \caption{A sample of the SAE/AAVE synonym mapping that we learned automatically from corpus data.}
    \label{tab:lexical_mapping}
\end{table}

\subsection{Transformed Datasets}
\begin{table*}[]
    \centering
    \resizebox{\textwidth}{!}{%
    \begin{tabular}{lrrrrrrrrrrr}
    \toprule
    \textbf{Dataset} & \textbf{\# data} & \textbf{ass} & \textbf{aux} & \textbf{been} & \textbf{dey/it} & \textbf{got} & \textbf{lexical} & \textbf{neg cncrd}  & \textbf{null gen} & \textbf{null relcl} & \textbf{uninflect}\\ \midrule
CoLA&1,063& 9\%&  15\%& 6\%&  2\%&   2\%&   51\%&  4\%&    3\%&  3\%&   17\% \\
MNLI&9,682& 30\%& 20\%&  9\%&  4\%&  5\%&   69\%&  4\%&    11\%&   10\%&   23\% \\
QNLI&5,725& 16\%& 42\%&   2\%&  1\%&  3\%&   50\%&  1\%&    10\%&  4\%&   17\% \\
 QQP&390,690& 16\%&  2\%&  3\%&   63\%&  3\%&   59\%&  1\%&  3\%&  3\%&   13\% \\
RTE&3,029& 48\%&  40\%& 36\%&  3\%&   5\%&   81\%&  4\%&    28\%&   25&   40\% \\
SST-2&1,821& 31\%&  25\%& 5\%&  3\%&  4\%&   64\%&  4\%&    14\%&   15\%&   39\% \\
STS-B&1,894& 1\%&  $\sim$0&   32\%&  2\%& 3\%&  2\%&  9\%&    4\%&  2\%&  5\% \\
WNLI&146& 48\%&   36\%&  38\%&  3\%&   16\%&   90\%&  1\%&    37\%&   12\%&   33\%\\
    \bottomrule
    \end{tabular}
    }
    \caption{\textbf{Dataset statistics} reveal important differences between \data{} datasets, which come in markedly different sizes. The \textbf{\%} columns reflect the proportion of data points in which the primary sentence or question was modified using the given transformation (e.g. the existential \textit{dey/it}).
    }
    \label{tab:dataset_statistics}
\end{table*}

Our transformed tasks are all derived from GLUE. We skip \textit{Diagnostics} because it is not a benchmark, and we do not transform the Microsoft Research Paraphrase Corpus \cite{dolan2005automatically} because it is proprietary. However, we do transform the remaining seven benchmarks, which include the \textit{single-sentence tasks} (i) Stanford Sentiment Treebank (SST-2) which involves classifying the sentiment of movie reviews as positive or negative, and (ii) 
Corpus of Linguistic Acceptability (CoLA) which involves deciding whether a sentence is linguistically acceptable or not; the \textit{similarity and paraphrase task} called 
Semantic Textual Similarity Benchmark (STS-B), which involves predicting the similarity ratings between two sentences; and the \textit{inference tasks} (i) Multi-Genre Natural Language Inference (MNLI) which involves classifying the relationships between two sentences as entailment, contradiction, or neutral, (ii)  Question Natural Language Inference (QNLI) which involves predicting whether a given sentence is the correct answer to a given question; and finally (iii) Recognizing Textual Entailment (RTE) which involves predicting an entailment relation between two sentences. Table~\ref{tab:dataset_statistics} provides a set of summary statistics for these datasets. It is clear that they come in different sizes, and that the some tasks have been more heavily modified than others. However, most of the sentences in this benchmark have undergone at least one transformation.

\section{Speaker Validation and Gold-Standard}
\subsection{Validating Transformation Rules}
\label{sec:validating_tranformation_rules}
Since our morphosyntactic transformations are rule-based rather than data-driven, it is especially important to validate that these rules are aligned with real AAVE speakers' grammaticality judgments. 

\paragraph{User-Centered Validation Protocol.}
We opt for a participatory design process \cite{schuler1993participatory} to help ensure that these transformations are usable and meet the language practices of real speakers. 
We partnered with \org{,} \footnote{https://dataworkforce.gatech.edu/} an initiative started in Georgia Tech's College of Computing 
that seeks to involve members of underrepresented and economically disadvantaged groups in research and data annotation. All annotators were AAVE speakers and members of the Black community in Atlanta, and they were compensated for their time. Four volunteers from \org{} partnered in the design of this rule-validation process. Specifically, we co-designed appropriate questions to measure the linguistic and social plausibility of our transformation system. 

The HIT questions were based on a pair of utterances: (1) the original SAE sentence from the GLUE benchmark, and (2) the transformed AAVE sentence using only the morphosyntactic rules. We highlighted and indexed the portions of utterance (1) that were transformed in utterance (2), and we asked annotators for a binary grammaticality judgment.
Separately, we asked for the \textit{social} acceptability using a scale that was co-designed by \workers{}. Then, for text marked as ungrammatical, annotators provided us with the indices at which transformation errors occurred. The task was hosted on the Amazon Mechanical Turk sandbox platform, but we interfaced with the annotators throughout the entire annotation process to answer any questions.

In early iterations of the task, \workers{} discussed confusions and disagreements with the authors, and we discovered that the greatest variation in their judgments came not from differences in the speakers' underlying grammars, but rather from their different intuitions about what is \textit{socially} acceptable (alternatively awkward and unnatural) to say in certain social settings. To disentangle these factors, \workers{} helped us design a 10-point \textit{social acceptability} Likert scale with the following vernacular: \textit{If someone said this in your community, would it be (1) not very cool, (5) a bit sensitive, (7) passing, or (10) cool?} 

Separately, we discussed certain orthographic conventions that we had adopted from the linguistics literature. \workers{} indicated that some of these conventions were disagreeable -- especially the spelling for \textit{there are} as \textit{dey} from \citet{green2002african}. Some \workers{} suggested we use the spelling \textit{dey're} instead. Relatedly, the \workers{} found the \textit{ass constructions} sensitive, given its long history of mischaracterization as an expletive, as well as the broader relationship between such dialect misunderstandings and racial injustice \cite{rickford2016language,rickford2016raciolinguistics}. We simply excluded \textit{ass constructions} from the validation. \workers{} also reported sentences from the original GLUE task were highly offensive (e.g. mentions of sexual violence). We used the Perspective API\footnote{\url{https://www.perspectiveapi.com/}} and the offensive language classifier \citet{davidson2017automated} to filter out such instances.

\begin{table}
\centering
\resizebox{\columnwidth}{!}{%
\def\arraystretch{1.15}
\begin{tabular}{lccc}\toprule

 & \textbf{Accuracy} & \textbf{Accuracy} & \textbf{Size} \\
 \textbf{Transformation} & (Maj. Vote) & (Unanimous) & \textbf{$n$} \\ \midrule
 
    \textit{Ass} constructions & - & - & - \\
Auxiliaries&96.6&77.4&638\\
Been / done&95.4&72.7&670\\
Existential dey/it&91.4&57.9&304\\
Gonna / finna&95.4&78.7&197\\
Have / got&96.2&84.8&290\\
Inflection&97.1&82.3&761\\
Negative concord&95.9&73.6&584\\
Negative inversion&95.0&69.3&101\\
Null genitives&97.9&85.3&573\\
Relative clause structures&94.1&58.3&489\\
\bottomrule 
\end{tabular}
}
\caption{Accuracy of SAE$\rightarrow$AAVE transformations and $n$ the number of instances present.
}
\label{tab:dataworks}
\end{table}

Finally, we discussed the visual and interactive elements of the task itself. Workers preferred to see the synthetic AAVE text appear with visual priority above the SAE sentence. We also adjusted the color scheme to maximally distinguish concepts of social and grammatical acceptability. The word \textit{acceptability} itself was triggering for the \workers{} because it evoked the history of linguistic discrimination against AAVE speakers based on ignorant and prescriptive claims regarding ``correct'' or ``proper'' English. For this reason, we modified the prompt to read: \textit{Do the words and the order of the words make sense?} With extensive follow-up meetings, we clarified that \textit{to make sense} means a sentence follows expected and consistent language rules (i.e. a speaker's internal grammar). 

\vspace{0.08in}
\noindent
\textbf{Results.}
In the end, we collected acceptability judgments from three \org{} workers for each of 2,556 randomly sampled sentence transformation pairs. We observed fair inter-annotator agreement with Krippendorf's $\alpha$=$0.26$. Table~\ref{tab:dataworks} presents the aggregate judgments for local transformations in each 
morphological category. 
Here, we report the transformation accuracy as the proportion of local transformations marked as acceptable by majority vote or unanimous consensus, and we find our transformation rules are strongly validated. Majority vote gives nearly \textbf{100\% accuracy} for all transformation types.
Even under \emph{strict} unanimous consensus, the accuracy exceeds 70\% for seven of the 11 transformation types. 
Overall, this shows the quality of our linguistic transformations.

\vspace{0.04in}
\noindent
\textbf{Error Analysis.} Although our transformation rules are generally valid, errors can stem from an overapplication of the rule in restricted contexts. For example, most rules do not apply to idioms or named entities, so if we see a brand name like \textit{Reese's Pieces}, we should not remove the possessive \textit{s}. 
Other observed challenge cases include the subjunctive mood and subject inversions in questions, the non-standard morphology of certain contractions, as well as co-reference and scoping issues in relative clauses, ellipsis, and long-range dependencies (See Appendix~\ref{appdx:detailed_error_analysis} for more details). These each may introduce their own special cases that could be coded in future iterations. For a more reliable test set, we next construct a gold standard in Section~\ref{subsec:gold_test_set}

\subsection{Building a Gold Test Set}
\label{subsec:gold_test_set}

\begin{table*}[]
    \centering
    \resizebox{\textwidth}{!}{%
    \begin{tabular}{rrrrrrrr}
    \toprule
    & \textbf{MNLI} & \textbf{QNLI} & \textbf{QQP} & \textbf{RTE} & \textbf{SST-2} & \textbf{STS-B} & \textbf{WNLI}\\ \midrule
    \# Hand-Validated Synthetic Transformations  & 3,489 &    2,690 &    2,932 &   547 &    404 &   1,608 &   225 \\
    \# Natural AAVE Sentence Translations & 930 &     633 &    2,696 &   182 &    188 &    474 &   204 \\\midrule
    \textbf{Gold Test Set} Total Size & 4,419 & 3,323 & 5,628 & 792 & 592 & 2,082 & 429 \\
    \bottomrule
    \end{tabular}
    }
    \caption{\textbf{Gold Test Set} size for each NLU task. 
    }
    \label{tab:gold_annotations}
\end{table*}

Despite the advantages of controllable feature transformations for benchmarking with explainable error analysis, we cannot rely on the synthetic benchmark alone. Synthetic data may not fully capture the social and structural nuances of AAVE, nor speakers' dynamic and contextual use of dialect feature density. This motivates us to build a small test set of Gold Standard AAVE utterances. Here, annotators considered GLUE sentence transformations as before. The \workers{} could either (1) confirm that synthetic transformation was natural, or alternatively (2) provide us with their own \textit{translation} of the SAE text. Together, datapoints from (1) and (2) construct our Gold Test Set.\footnote{\textbf{Note:} we did not build a Gold CoLA Test set. The nature of the annotation task would be ambiguous since CoLA itself contains intentionally ungrammatical utterances. It is not clear how annotators should translate ungrammatical SAE into ungrammatical AAVE.} We provide the distribution of Gold Standard datapoints for each task in Table~\ref{tab:gold_annotations}. In future iterations, we will expand the total size of the Gold Test sets for reliable benchmarking.

\section{Benchmarking Models on \data{}}

In this section, we stress-test current systems on NLU tasks and reveal performance drops on dialect-variants. We investigate the effectiveness of standard training on \data{} and we ablate the dialect test set to understand which dialect features most significantly challenge models.

We have two variants of synthetic AAVE data. In \textbf{AAVE (\data{})}, we apply the full suite of Section~\ref{sec:data} transformations to the standard GLUE tasks. In \textbf{AAVE Morph}, we have an ablated variant of \data{} where only the morphosyntactic transformations (Section~\ref{subsec:morphosyntactic_translation}) are executed. By testing base SAE models on this data, we can disentangle the challenges associated with vocabulary shift from those associated with structural differences. If the challenges of \data{} were entirely lexical, we would anticipate that any performance disparity could be recovered with domain-specific word embeddings, since prior work has found such embeddings adequately represent the meanings of new words in AAVE corpora \cite{hwang2020towards}.

\subsection{Standard Training}
The most direct way to prepare models for a particular language variety is to directly train them on a dialect-variant of the task. Using our transformation rules (Section~\ref{sec:data}), we first augment the GLUE training set with AAVE features and then re-train the models (125M-parameter RoBERTA-base) on the augmented data. Following \citet{liu2019roberta},  the batch size was 16. The maximum learning rate was selected as $5e-4$ and the maximum number of training epochs was set to be either $5$ or $10$.

\subsection{Results}

Table~\ref{tab:performance} compares the performance of RoBERTa models trained and tested on SAE or AAVE-variants of seven natural language understanding tasks in GLUE. Results are given as Matthew's Correlation for CoLA, Pearson-Spearman Correlation for STS-B, and Accuracy for all other tasks, averaged over three random seeds. In most cases, training jointly on GLUE and VALUE (SAE + AAVE) leads to \textbf{best performance}. With a single training set, there is an expected \best{pattern:} training with the corresponding train set typically leads to best performance on the corresponding test set. With the exception of RTE,\footnote{RTE may be an outlier because of variance due to its small size: only 2.5k data points vs. QNLI with 100k} base models all suffer a \drop{drop} in performance when tested on the full AAVE (\data{}) test set compared with the models trained on AAVE or jointly on SAE + AAVE (e.g., a 1.5\% drop on SST-2; a 0.9\% drop on QNLI compared to SAE + AAVE). Performance gaps of a similar magnitude are observed when we test on the Gold Test set (e.g., a 1.2\% drop on SST-2; a 0.8\% drop on QNLI). Further effort is needed to make the current NLU models more robust to dialect variations.

We also see that AAVE Morph challenges current models, which suggests that strategies for resolving any performance gap should take dialect morphology and syntax into consideration. Compared to the AAVE column, there is a less severe but still visible drop in AAVE Morph testing: from 94.3 to 93.2 in SST-2, and from 92.6 to 92.0 in QNLI, for instance. 
Thus we conclude that the challenge with dialects extends beyond a mere difference in the lexicon.

\begin{table}[h!]
\resizebox{\columnwidth}{!}{%
\def\arraystretch{1.15}
\begin{tabular}{clc|cc|c}\toprule
 &&Test&\multicolumn{2}{c|}{Synth. Testing} & Gold
 \\ 
 & Training & SAE & Morph. & AAVE & Test \\[1.5mm] \midrule
 
 \parbox[t]{2mm}{\multirow{4}{*}{\rotatebox[origin=c]{90}{\textbf{CoLA}}}} & SAE (GLUE) & \best{56.3} & 55.7 & \drop{55.6} & -\\ 
 & AAVE Morph. & 56.3 & \best{56.0} & 55.4 & - \\
 & AAVE (\data) & 56.2 & 55.6 & \best{55.8} &  -\\
 & SAE + AAVE & \textbf{57.8} & \textbf{56.2} & \textbf{56.5} & -\\
 \midrule
 
 \parbox[t]{2mm}{\multirow{4}{*}{\rotatebox[origin=c]{90}{\textbf{MNLI}}}} & SAE (GLUE) & \best{83.6} & 83.0 & \drop{82.8}& \drop{82.1}\\
 & AAVE Morph. & 82.2& \best{83.3} & 82.5 & 82.3\\
 & AAVE (\data) & 83.1 & 83.2 & \best{83.5} & \best{82.9}\\
 & SAE + AAVE & \textbf{83.8} & \textbf{83.6} & \textbf{83.6} & \textbf{83.3}\\
 \midrule
 
 \parbox[t]{2mm}{\multirow{4}{*}{\rotatebox[origin=c]{90}{\textbf{QNLI}}}} & SAE (GLUE) & \best{\textbf{92.8}} & 92.0 & \drop{91.4} & \drop{91.2} \\
 & AAVE Morph. & 92.5 & \best{\textbf{92.6}} & 91.2  & 91.2 \\
 & AAVE (\data) & 92.5 & 92.4 & \best{91.8} & \best{91.8}\\
 & SAE + AAVE & \textbf{92.8}& 92.5 & \textbf{92.3} & \textbf{92.0}\\
 \midrule
 
 \parbox[t]{2mm}{\multirow{4}{*}{\rotatebox[origin=c]{90}{\textbf{RTE}}}} & SAE (GLUE) & 66.4 & 66.4 & 67.8 & 67.6 \\
 & AAVE Morph. & \best{\textbf{68.9}} & \best{68.2}	&\best{\textbf{69.7}} & \best{\textbf{68.8}} \\
 & AAVE (\data) &67.1 &  66.1&67.2& 67.3\\
 & SAE + AAVE & 68.6 & \textbf{68.9} & 69.1	& 67.6\\
 \midrule
 
 \parbox[t]{2mm}{\multirow{4}{*}{\rotatebox[origin=c]{90}{\textbf{SST-2}}}} & SAE (GLUE) & \best{94.6} & 93.2 & \drop{92.4} & \drop{92.0}\\
 & AAVE Morph. &  94.0	& \best{94.3} & 92.3 & 92.1\\
 & AAVE (\data)  & 94.0 & 93.8 	 & \best{93.0}	& \best{92.8}\\
 & SAE + AAVE & \textbf{94.8} &  \textbf{94.5} & \textbf{93.9} & \textbf{93.2} \\
 \midrule
 
 \parbox[t]{2mm}{\multirow{4}{*}{\rotatebox[origin=c]{90}{\textbf{STS-B}}}} & SAE (GLUE) & \best{\textbf{89.4}}	& 88.3 & \best{88.5} & \drop{88.2}\\
 & AAVE Morph. & 89.1 & \best{88.9} & 88.0 & 88.2  \\
 & AAVE (\data) & 88.8& 88.7	& 88.3 & \best{88.3}\\
 & SAE + AAVE & 89.2 & \textbf{89.0} & \textbf{88.9}  & \textbf{88.5} \\
 \hline
 
 \parbox[t]{2mm}{\multirow{4}{*}{\rotatebox[origin=c]{90}{\textbf{QQP}}}} & SAE (GLUE) & \best{\textbf{90.9}}& 89.8 & \drop{89.5} & \drop{89.2}\\
 & AAVE Morph. & 90.1 & \best{90.2} & 89.6  & 89.3\\
 & AAVE (\data) & 90.3  & 89.6 & \best{89.8} & \best{89.6}\\
 & SAE + AAVE & 90.8 & \textbf{90.3} & \textbf{90.1} & \textbf{89.8} \\
\bottomrule 
\end{tabular}
}
\caption{\textbf{Dialect understanding results} for six tasks (Matthew's Corr. for CoLA; Pearson-Spearman Corr. for STS-B; Accuracy for all others). AAVE Morph is a subset of \data{} in which only the morphosyntactic transformations (Section~\ref{subsec:morphosyntactic_translation}) are executed. SAE + AAVE indicates training on the merged GLUE and VALUE train sets. Best performance is given in \textbf{bold} and best performance with a single train set is given in \best{gray.} The result gaps are significant. With the exception of RTE and STS-B, models trained on SAE (GLUE) suffer a \drop{drop} in performance when tested on synthetic AAVE (\data{}) or the Gold Test.
}
\label{tab:performance}
\end{table}

\begin{table*}[h!]
\centering
\resizebox{0.80\textwidth}{!}{%
\def\arraystretch{1.15}
\begin{tabular}{lr|rrrrrr|r}
\toprule
Feature &     $r_T$ &      \textbf{c}$\rightarrow$\textbf{n} &      \textbf{c}$\rightarrow$\textbf{e} &      \textbf{n}$\rightarrow$\textbf{c} &      \textbf{n}$\rightarrow$\textbf{e} &      \textbf{e}$\rightarrow$\textbf{c} &      \textbf{e}$\rightarrow$\textbf{n} & $|\mathcal{X}_T|$\\
\midrule
Auxiliaries                &  4.20 &  0.20 &  0.07 &  \textbf{1.62} &  0.88 &  0.68 &  0.74 & 1,477 \\
Been / done                &  3.06 &  0.22 &  0.00 &  \textbf{1.31} &  0.44 &  0.22 &  0.88 & 457 \\
Inflection                 &  2.88 &  0.33 &  0.20 &  0.59 &  0.46 &  0.39 &  \textbf{0.92} &  1,526 \\
Lexical                    &  5.92 & 0.67 &  0.27 &  1.35 &  0.57 &  0.88 &  \textbf{2.18} &  4,902 \\
Negative concord           &  6.88 &  0.64 &  0.16 &  \textbf{2.56} &  0.16 &  2.08 &  1.28 &   625 \\
Negative inversion         &  \textbf{9.09} &  0.00 &  0.00 &  0.00 &  \textbf{9.09} &  0.00 &  0.00 &    11 \\
Relative clause structures &  5.86 &  0.31 &  0.62 &  1.23 &  0.62 &  0.31 &  \textbf{2.78} &   324 \\
\bottomrule
\end{tabular}
}
\caption{\small{\textbf{Perturbation analysis.} The first column $r_T$ gives the proportion of testing instances where the introduction of a particular dialect feature results in a new model error. This column indicates that \textit{negative inversions} are the most challenging for MNLI. The final column gives the size of the set $\mathcal{X}_T$, which is the denominator in the ratio $r_T$. The remaining columns indicate the contributions of different error types to the cumulative $r_T$: the model flips the correct label on the left $\rightarrow$ into the incorrect label on the right side. \textbf{c}: contradiction; \textbf{n}: neutral; \textbf{e}: entailment.}}
\label{tab:transform_causes_error}
\end{table*}
\subsection{Perturbation Analysis}

Finally, we run a perturbation analysis \cite{alvarez2017causal} to better understand the impact of each dialectal feature on model performance. For the sake of simplicity, we focus only on MNLI. Specifically, we are interested in cases where the introduction of a particular feature results in a model error. Therefore, we count, for each feature transformation function $T$, the number of sentence pairs $(\boldsymbol{x}^0_i, \boldsymbol{x}^1_i)$ for which a GLUE-trained RoBERTA model $f$ changes its prediction from a correct inference $y_i$ to an incorrect inference under the transformation. Not all sentence structures allow for new features, so we consider only the subset of pairs for which the transformation is effective in the hypothesis sentence, and where the original GLUE pair had been predicted correctly.
Then the ratio $r_T$ is be defined as: 
\begin{equation*}
    r_T = \frac{\left|\left\{(\boldsymbol{x}^0_i, \boldsymbol{x}^1_i) \in \mathcal{X}_T \text{ : } f(\boldsymbol{x}^0_i, T(\boldsymbol{x}^1_i)) \not= y_i \right|  \right\}}{|\mathcal{X}_T|}
\end{equation*}
Here $\mathcal{X}_T$ is: 
\begin{align*}
    \mathcal{X}_T &= \{(\boldsymbol{x}^0_i, \boldsymbol{x}^1_i) \text{ : } T(\boldsymbol{x}^1_i) \not= \boldsymbol{x}^1_i \land f(\boldsymbol{x}^0_i, \boldsymbol{x}^1_i) = y_i\}
\end{align*}
and $r_T$ indicates the proportion of inferences that were flipped to an incorrect label in the presence of $T$. We report this ratio for each feature  in Table~\ref{tab:transform_causes_error}.

The first column in table shows that, when we introduce a \textit{negative inversion} into a Hypothesis sentence for which the GLUE-trained RoBERTa model was originally correct, then in 9.09\% of cases, that correct label would be flipped to an incorrect one.\footnote{This is the highest error rate for any transformation rule. Note that $|\mathcal{X}_T| = 11$ datapoints is a much smaller sample size so the $r_T$ estimate is more variable.} The \textit{inflection} rule and \textit{been / done} constructions appear less challenging, but still result in 2.88\% and 3.06\% of new errors respectively. 
The remaining table columns indicate the contributions of different model mistakes to the overall $r_T$ ratio. For example, the single error due to negative inversion occurs here when the model mistakes a neutral relationship for entailment (n$\rightarrow$e) in the following pair: 
\textsc{Premise}: ``\emph{Still, commercial calculation isn't sufficient to explain his stand}'' and \textsc{Hypothesis}: ``\textit{Won't nothing be enough to explain his strong opinion}''. In negative concord environments, we most often see neutral pairs mistakenly labeled as contradictory (n$\rightarrow$c), as with the \textsc{Premise}: ``\emph{Each state is different...}'' and \textsc{Hypothesis}: ``\textit{You can go from one area of a state to another and not see no resemblance}. For more examples, see Tables \ref{tab:perturbation_error_examples_1} and \ref{tab:perturbation_error_examples_2} in Appendix~\ref{appdx:perturbation_analysis}.

\vspace{-0.04in}
\section{Why Not Use Style Transfer?}\vspace{-0.04in}
We qualitatively investigated the differences between our rule-based approach and a very well-performing unsupervised dialect style transfer model, \textsc{strap} \citet{krishna2020reformulating}. To train \textsc{strap}, we created a pseudo-parallel corpus using a diverse paraphrase model to paraphrase different styles of text, including SAE and the AAVE text from the \texttt{TwitterAAE} corpus \citet{blodgett2018twitter}. Then we fine-tuned a GPT-2 model as the inverse paraphrase function, which learned to reconstruct the various styles. We used the SAE paraphrase model and the AAVE inverse paraphrase model to transfer from SAE to AAVE.
In general, we found that STRAP is capable of much greater output diversity. However, in a systematic analysis of dialectal NLU, the first goal is to ensure that the underlying relationships like \textit{entailment} are not distorted. STRAP can distort the meaning of the text with hallucinations and deletion of key details. Our transformation approach preserves the meaning of the text and thus better captures AAVE morphosyntax. See Appendix~\ref{appdx:qualitative_evalue_strap} for more details. 
\vspace{-0.04in}
\section{Conclusion}\vspace{-0.04in}
This work introduces the English VernAcular Language Understanding Evaluation (\data) benchmark, a challenging variant of GLUE that we created with a set of lexical and morphosyntactic transformation rules. 
We constructed rules for \numAAVEFeatures{} features of AAVE, and recruit fluent AAVE speakers to validate each feature transformation via linguistic acceptability judgments in a participatory design manner.
Experiments show that the introduction of new dialectal features can lead to a drop in performance.  We also test methods for efficiently adapting models to different language varieties, and discuss dialect specific challenges that our current NLP models are struggling with. 
Our work sheds light on the disparities of language technologies and has key implications for facilitating more dialect-competent NLU systems.
Our longer term goals are to expand \data{} to more NLP tasks such as CoQA \cite{reddy2019coqa}, and to include other dialects such as Indian English \cite{demszky2020learning,lange2012syntax,bhatt2008indian} and Singapore English \cite{wee2008singapore}.

\paragraph{Limitations and Considerations.} Researchers and practitioners should keep the following limitations and considerations in mind when using \data{}. Firstly, 
dialects are not the deterministic speech patterns that our transformation rules might suggest. While speakers of a dialect have \textit{linguistic competence} over systematic and internalized grammar rules, speakers still posses an individual degree of control over which features they will employ \cite{coupland2007style}. The density of these features can vary, not only along demographic axes of geography, age, and gender \cite{nguyen2016computational}, but also with different identity presentations in different social contexts \cite{bucholtz2005identity}. We use \data{} to stress-test current systems by maximally modifying current resources with feature transformations. The high density of dialectal features may appear exaggerated here. Secondly, linguists have historically studied dialects through oral speech via live interviews \cite{rickford2002linguists}. The descriptions of academic references will not always map perfectly to the written domain (see Section~\ref{sec:validating_tranformation_rules} on the spelling of \textit{dey}). The orthographic conventions of language communities may vary as significantly as do speech patterns. A third and critical concern is the limitation of synthetic data. Synthetic transformations have the advantage of allowing carefully controlled perturbation analysis and scaling up this analysis without the expensive creation of new datasets. However, synthetic data will not fully capture the social and structural nuances of AAVE, nor speakers' dynamic and contextual use of dialect feature density. For this reason, it is important to ultimately test user-facing models on domain-specific and gold-standard dialectal data. We are continuing to expand our gold-standard test set for GLUE tasks. A fourth consideration is the history of linguistic discrimination and the broader relationship between such dialect misunderstandings and racial injustice \cite{rickford2016language,rickford2016raciolinguistics}. AAVE has been frequently appropriated and misused by non-Black individuals, especially in online contexts \cite{reyes2005appropriation, ilbury2020sassy}. To mitigate deployment risks, we ask users to sign a Data Use Agreement (See Ethics Section).

\section*{Acknowledgements} The authors would like to thank the \org{} team, as well as Elizabeth DiSalvo, Rahual Gupta, and Jwala Dhamala for their helpful feedback. CZ is supported by the NSF Graduate Research Fellowship under Grant No. DGE-2039655. DY is supported by the Microsoft Research Faculty Fellowship. 
This work is 
funded in part by Amazon Research Award under the Alexa Fairness in AI. 
\section*{Ethics}
Our task comes from the public version of GLUE \cite{wang2018glue}. Our annotation efforts revealed non-normative and offensive language in these original datasets, and we caution practitioners to be aware of this.
The rules for converting SAE to AAVE are linguistically informed, and are not designed to change the original meaning of the sentence.
Due to the participatory design nature of this work, we involved AAVE speakers and volunteers in the task creation and rule validation process. We asked annotators to skip a specific task and take a break if they are overwhelmed with the task. 
Our annotators were compensated by \org{} for their time, and volunteered to help build this linguistic resource for their dialects. 
Note that AAVE is spoken, and our work only involves speakers from Atlanta.
We ask that all users sign the following online agreement before using this resource: ``\textit{I will not use \data{} for malicious purposes including (but not limited to): deception, impersonation, mockery, discrimination, hate speech, targeted harassment and cultural appropriation. In my use of this resource, I will respect the dignity and privacy of all people.}''

\bibliographystyle{acl_natbib}

\begin{thebibliography}{73}
\expandafter\ifx\csname natexlab\endcsname\relax\def\natexlab#1{#1}\fi

\bibitem[{Alvarez-Melis and Jaakkola(2017)}]{alvarez2017causal}
David Alvarez-Melis and Tommi Jaakkola. 2017.
\newblock \href {https://doi.org/10.18653/v1/D17-1042} {A causal framework for
  explaining the predictions of black-box sequence-to-sequence models}.
\newblock In \emph{Proceedings of the 2017 Conference on Empirical Methods in
  Natural Language Processing}, pages 412--421, Copenhagen, Denmark.
  Association for Computational Linguistics.

\bibitem[{Alzantot et~al.(2018)Alzantot, Sharma, Elgohary, Ho, Srivastava, and
  Chang}]{alzantot2018generating}
Moustafa Alzantot, Yash Sharma, Ahmed Elgohary, Bo-Jhang Ho, Mani Srivastava,
  and Kai-Wei Chang. 2018.
\newblock \href {https://doi.org/10.18653/v1/D18-1316} {Generating natural
  language adversarial examples}.
\newblock In \emph{Proceedings of the 2018 Conference on Empirical Methods in
  Natural Language Processing}, pages 2890--2896, Brussels, Belgium.
  Association for Computational Linguistics.

\bibitem[{Belinkov and Bisk(2018)}]{belinkov2017synthetic}
Yonatan Belinkov and Yonatan Bisk. 2018.
\newblock \href {https://openreview.net/forum?id=BJ8vJebC-} {Synthetic and
  natural noise both break neural machine translation}.
\newblock In \emph{6th International Conference on Learning Representations,
  {ICLR} 2018, Vancouver, BC, Canada, April 30 - May 3, 2018, Conference Track
  Proceedings}. OpenReview.net.

\bibitem[{Bender(2000)}]{bender2000syntactic}
Emily~M Bender. 2000.
\newblock \emph{Syntactic variation and linguistic competence: The case of AAVE
  copula absence}.
\newblock stanford university Stanford, California.

\bibitem[{Bhatt(2008)}]{bhatt2008indian}
Rakesh~M Bhatt. 2008.
\newblock Indian english: Syntax.
\newblock In \emph{A handbook of varieties of English}, pages 2208--2222. De
  Gruyter Mouton.

\bibitem[{Blodgett et~al.(2016)Blodgett, Green, and
  O{'}Connor}]{blodgett2016demographic}
Su~Lin Blodgett, Lisa Green, and Brendan O{'}Connor. 2016.
\newblock \href {https://doi.org/10.18653/v1/D16-1120} {Demographic dialectal
  variation in social media: A case study of {A}frican-{A}merican {E}nglish}.
\newblock In \emph{Proceedings of the 2016 Conference on Empirical Methods in
  Natural Language Processing}, pages 1119--1130, Austin, Texas. Association
  for Computational Linguistics.

\bibitem[{Blodgett and O'Connor(2017)}]{DBLP:journals/corr/BlodgettO17}
Su~Lin Blodgett and Brendan O'Connor. 2017.
\newblock \href {http://arxiv.org/abs/1707.00061} {Racial disparity in natural
  language processing: {A} case study of social media african-american
  english}.
\newblock \emph{CoRR}, abs/1707.00061.

\bibitem[{Blodgett et~al.(2018)Blodgett, Wei, and
  O{'}Connor}]{blodgett2018twitter}
Su~Lin Blodgett, Johnny Wei, and Brendan O{'}Connor. 2018.
\newblock \href {https://doi.org/10.18653/v1/P18-1131} {{T}witter {U}niversal
  {D}ependency parsing for {A}frican-{A}merican and mainstream {A}merican
  {E}nglish}.
\newblock In \emph{Proceedings of the 56th Annual Meeting of the Association
  for Computational Linguistics (Volume 1: Long Papers)}, pages 1415--1425,
  Melbourne, Australia. Association for Computational Linguistics.

\bibitem[{Bolukbasi et~al.(2016)Bolukbasi, Chang, Zou, Saligrama, and
  Kalai}]{bolukbasi2016man}
Tolga Bolukbasi, Kai{-}Wei Chang, James~Y. Zou, Venkatesh Saligrama, and
  Adam~Tauman Kalai. 2016.
\newblock \href
  {https://proceedings.neurips.cc/paper/2016/hash/a486cd07e4ac3d270571622f4f316ec5-Abstract.html}
  {Man is to computer programmer as woman is to homemaker? debiasing word
  embeddings}.
\newblock In \emph{Advances in Neural Information Processing Systems 29: Annual
  Conference on Neural Information Processing Systems 2016, December 5-10,
  2016, Barcelona, Spain}, pages 4349--4357.

\bibitem[{Bowman et~al.(2015)Bowman, Angeli, Potts, and
  Manning}]{bowman2015large}
Samuel~R. Bowman, Gabor Angeli, Christopher Potts, and Christopher~D. Manning.
  2015.
\newblock \href {https://doi.org/10.18653/v1/D15-1075} {A large annotated
  corpus for learning natural language inference}.
\newblock In \emph{Proceedings of the 2015 Conference on Empirical Methods in
  Natural Language Processing}, pages 632--642, Lisbon, Portugal. Association
  for Computational Linguistics.

\bibitem[{Bucholtz and Hall(2005)}]{bucholtz2005identity}
Mary Bucholtz and Kira Hall. 2005.
\newblock Identity and interaction: A sociocultural linguistic approach.
\newblock \emph{Discourse studies}, 7(4-5):585--614.

\bibitem[{Collins et~al.(2008)Collins, Moody, and Postal}]{collins2008aae}
Chris Collins, Simanique Moody, and Paul~M Postal. 2008.
\newblock An aae camouflage construction.
\newblock \emph{Language}, pages 29--68.

\bibitem[{Coupland(2007)}]{coupland2007style}
Nikolas Coupland. 2007.
\newblock \emph{Style: Language variation and identity}.
\newblock Cambridge University Press.

\bibitem[{Davidson et~al.(2017)Davidson, Warmsley, Macy, and
  Weber}]{davidson2017automated}
Thomas Davidson, Dana Warmsley, Michael Macy, and Ingmar Weber. 2017.
\newblock \href {https://arxiv.org/abs/1703.04009} {Automated hate speech
  detection and the problem of offensive language}.
\newblock \emph{ArXiv preprint}, abs/1703.04009.

\bibitem[{Demszky et~al.(2021{\natexlab{a}})Demszky, Sharma, Clark,
  Prabhakaran, and Eisenstein}]{demszky-etal-2021-learning}
Dorottya Demszky, Devyani Sharma, Jonathan Clark, Vinodkumar Prabhakaran, and
  Jacob Eisenstein. 2021{\natexlab{a}}.
\newblock \href {https://doi.org/10.18653/v1/2021.naacl-main.184} {Learning to
  recognize dialect features}.
\newblock In \emph{Proceedings of the 2021 Conference of the North American
  Chapter of the Association for Computational Linguistics: Human Language
  Technologies}, pages 2315--2338, Online. Association for Computational
  Linguistics.

\bibitem[{Demszky et~al.(2021{\natexlab{b}})Demszky, Sharma, Clark,
  Prabhakaran, and Eisenstein}]{demszky2020learning}
Dorottya Demszky, Devyani Sharma, Jonathan Clark, Vinodkumar Prabhakaran, and
  Jacob Eisenstein. 2021{\natexlab{b}}.
\newblock \href {https://doi.org/10.18653/v1/2021.naacl-main.184} {Learning to
  recognize dialect features}.
\newblock In \emph{Proceedings of the 2021 Conference of the North American
  Chapter of the Association for Computational Linguistics: Human Language
  Technologies}, pages 2315--2338, Online. Association for Computational
  Linguistics.

\bibitem[{Dolan and Brockett(2005)}]{dolan2005automatically}
William~B. Dolan and Chris Brockett. 2005.
\newblock \href {https://aclanthology.org/I05-5002} {Automatically constructing
  a corpus of sentential paraphrases}.
\newblock In \emph{Proceedings of the Third International Workshop on
  Paraphrasing ({IWP}2005)}.

\bibitem[{Dorn(2019)}]{dorn2019dialect}
Rachel Dorn. 2019.
\newblock \href {https://doi.org/10.26615/issn.2603-2821.2019_003}
  {Dialect-specific models for automatic speech recognition of {A}frican
  {A}merican {V}ernacular {E}nglish}.
\newblock In \emph{Proceedings of the Student Research Workshop Associated with
  RANLP 2019}, pages 16--20, Varna, Bulgaria. INCOMA Ltd.

\bibitem[{Ebrahimi et~al.(2018)Ebrahimi, Rao, Lowd, and
  Dou}]{ebrahimi2018hotflip}
Javid Ebrahimi, Anyi Rao, Daniel Lowd, and Dejing Dou. 2018.
\newblock \href {https://doi.org/10.18653/v1/P18-2006} {{H}ot{F}lip: White-box
  adversarial examples for text classification}.
\newblock In \emph{Proceedings of the 56th Annual Meeting of the Association
  for Computational Linguistics (Volume 2: Short Papers)}, pages 31--36,
  Melbourne, Australia. Association for Computational Linguistics.

\bibitem[{Gardner et~al.(2020)Gardner, Artzi, Basmov, Berant, Bogin, Chen,
  Dasigi, Dua, Elazar, Gottumukkala, Gupta, Hajishirzi, Ilharco, Khashabi, Lin,
  Liu, Liu, Mulcaire, Ning, Singh, Smith, Subramanian, Tsarfaty, Wallace,
  Zhang, and Zhou}]{gardner-etal-2020-evaluating}
Matt Gardner, Yoav Artzi, Victoria Basmov, Jonathan Berant, Ben Bogin, Sihao
  Chen, Pradeep Dasigi, Dheeru Dua, Yanai Elazar, Ananth Gottumukkala, Nitish
  Gupta, Hannaneh Hajishirzi, Gabriel Ilharco, Daniel Khashabi, Kevin Lin,
  Jiangming Liu, Nelson~F. Liu, Phoebe Mulcaire, Qiang Ning, Sameer Singh,
  Noah~A. Smith, Sanjay Subramanian, Reut Tsarfaty, Eric Wallace, Ally Zhang,
  and Ben Zhou. 2020.
\newblock \href {https://doi.org/10.18653/v1/2020.findings-emnlp.117}
  {Evaluating models{'} local decision boundaries via contrast sets}.
\newblock In \emph{Findings of the Association for Computational Linguistics:
  EMNLP 2020}, pages 1307--1323, Online. Association for Computational
  Linguistics.

\bibitem[{Green(2002)}]{green2002african}
Lisa~J Green. 2002.
\newblock \emph{African American English: a linguistic introduction}.
\newblock Cambridge University Press.

\bibitem[{Groenwold et~al.(2020)Groenwold, Ou, Parekh, Honnavalli, Levy, Mirza,
  and Wang}]{groenwold2020investigating}
Sophie Groenwold, Lily Ou, Aesha Parekh, Samhita Honnavalli, Sharon Levy, Diba
  Mirza, and William~Yang Wang. 2020.
\newblock \href {https://doi.org/10.18653/v1/2020.emnlp-main.473}
  {Investigating {A}frican-{A}merican {V}ernacular {E}nglish in
  transformer-based text generation}.
\newblock In \emph{Proceedings of the 2020 Conference on Empirical Methods in
  Natural Language Processing (EMNLP)}, pages 5877--5883, Online. Association
  for Computational Linguistics.

\bibitem[{Halevy et~al.(2021)Halevy, Harris, Bruckman, Yang, and
  Howard}]{halevy2021mitigating}
Matan Halevy, Camille Harris, Amy Bruckman, Diyi Yang, and Ayanna Howard. 2021.
\newblock Mitigating racial biases in toxic language detection with an
  equity-based ensemble framework.
\newblock In \emph{Equity and Access in Algorithms, Mechanisms, and
  Optimization}, pages 1--11.

\bibitem[{Hovy and Spruit(2016)}]{hovy2016social}
Dirk Hovy and Shannon~L. Spruit. 2016.
\newblock \href {https://doi.org/10.18653/v1/P16-2096} {The social impact of
  natural language processing}.
\newblock In \emph{Proceedings of the 54th Annual Meeting of the Association
  for Computational Linguistics (Volume 2: Short Papers)}, pages 591--598,
  Berlin, Germany. Association for Computational Linguistics.

\bibitem[{Hwang et~al.(2020{\natexlab{a}})Hwang, Frey, and
  McKeown}]{hwang-etal-2020-towards}
Alyssa Hwang, William~R. Frey, and Kathleen McKeown. 2020{\natexlab{a}}.
\newblock \href {https://aclanthology.org/2020.vardial-1.15} {Towards
  augmenting lexical resources for slang and {A}frican {A}merican {E}nglish}.
\newblock In \emph{Proceedings of the 7th Workshop on NLP for Similar
  Languages, Varieties and Dialects}, pages 160--172, Barcelona, Spain
  (Online). International Committee on Computational Linguistics (ICCL).

\bibitem[{Hwang et~al.(2020{\natexlab{b}})Hwang, Frey, and
  McKeown}]{hwang2020towards}
Alyssa Hwang, William~R. Frey, and Kathleen McKeown. 2020{\natexlab{b}}.
\newblock \href {https://aclanthology.org/2020.vardial-1.15} {Towards
  augmenting lexical resources for slang and {A}frican {A}merican {E}nglish}.
\newblock In \emph{Proceedings of the 7th Workshop on NLP for Similar
  Languages, Varieties and Dialects}, pages 160--172, Barcelona, Spain
  (Online). International Committee on Computational Linguistics (ICCL).

\bibitem[{Ilbury(2020)}]{ilbury2020sassy}
Christian Ilbury. 2020.
\newblock “sassy queens”: Stylistic orthographic variation in twitter and
  the enregisterment of aave.
\newblock \emph{Journal of Sociolinguistics}, 24(2):245--264.

\bibitem[{Iyyer et~al.(2018)Iyyer, Wieting, Gimpel, and
  Zettlemoyer}]{iyyer2018adversarial}
Mohit Iyyer, John Wieting, Kevin Gimpel, and Luke Zettlemoyer. 2018.
\newblock \href {https://doi.org/10.18653/v1/N18-1170} {Adversarial example
  generation with syntactically controlled paraphrase networks}.
\newblock In \emph{Proceedings of the 2018 Conference of the North {A}merican
  Chapter of the Association for Computational Linguistics: Human Language
  Technologies, Volume 1 (Long Papers)}, pages 1875--1885, New Orleans,
  Louisiana. Association for Computational Linguistics.

\bibitem[{Jia and Liang(2017)}]{jia2017adversarial}
Robin Jia and Percy Liang. 2017.
\newblock \href {https://doi.org/10.18653/v1/D17-1215} {Adversarial examples
  for evaluating reading comprehension systems}.
\newblock In \emph{Proceedings of the 2017 Conference on Empirical Methods in
  Natural Language Processing}, pages 2021--2031, Copenhagen, Denmark.
  Association for Computational Linguistics.

\bibitem[{Jones et~al.(2020)Jones, Jia, Raghunathan, and
  Liang}]{jones2020robust}
Erik Jones, Robin Jia, Aditi Raghunathan, and Percy Liang. 2020.
\newblock \href {https://doi.org/10.18653/v1/2020.acl-main.245} {Robust
  encodings: A framework for combating adversarial typos}.
\newblock In \emph{Proceedings of the 58th Annual Meeting of the Association
  for Computational Linguistics}, pages 2752--2765, Online. Association for
  Computational Linguistics.

\bibitem[{J{\o}rgensen et~al.(2016)J{\o}rgensen, Hovy, and
  S{\o}gaard}]{jorgensen2016learning}
Anna J{\o}rgensen, Dirk Hovy, and Anders S{\o}gaard. 2016.
\newblock \href {https://doi.org/10.18653/v1/N16-1130} {Learning a {POS} tagger
  for {AAVE}-like language}.
\newblock In \emph{Proceedings of the 2016 Conference of the North {A}merican
  Chapter of the Association for Computational Linguistics: Human Language
  Technologies}, pages 1115--1120, San Diego, California. Association for
  Computational Linguistics.

\bibitem[{Jurgens et~al.(2017)Jurgens, Tsvetkov, and
  Jurafsky}]{jurgens2017incorporating}
David Jurgens, Yulia Tsvetkov, and Dan Jurafsky. 2017.
\newblock \href {https://doi.org/10.18653/v1/P17-2009} {Incorporating dialectal
  variability for socially equitable language identification}.
\newblock In \emph{Proceedings of the 55th Annual Meeting of the Association
  for Computational Linguistics (Volume 2: Short Papers)}, pages 51--57,
  Vancouver, Canada. Association for Computational Linguistics.

\bibitem[{Kaushik et~al.(2020)Kaushik, Hovy, and Lipton}]{Kaushik2020Learning}
Divyansh Kaushik, Eduard~H. Hovy, and Zachary~Chase Lipton. 2020.
\newblock \href {https://openreview.net/forum?id=Sklgs0NFvr} {Learning the
  difference that makes {A} difference with counterfactually-augmented data}.
\newblock In \emph{8th International Conference on Learning Representations,
  {ICLR} 2020, Addis Ababa, Ethiopia, April 26-30, 2020}. OpenReview.net.

\bibitem[{Kennedy(2007)}]{kennedy2007vagueness}
Christopher Kennedy. 2007.
\newblock Vagueness and grammar: The semantics of relative and absolute
  gradable adjectives.
\newblock \emph{Linguistics and philosophy}, 30(1):1--45.

\bibitem[{Kiritchenko and Mohammad(2018)}]{kiritchenko2018examining}
Svetlana Kiritchenko and Saif Mohammad. 2018.
\newblock \href {https://doi.org/10.18653/v1/S18-2005} {Examining gender and
  race bias in two hundred sentiment analysis systems}.
\newblock In \emph{Proceedings of the Seventh Joint Conference on Lexical and
  Computational Semantics}, pages 43--53, New Orleans, Louisiana. Association
  for Computational Linguistics.

\bibitem[{Koenecke et~al.(2020)Koenecke, Nam, Lake, Nudell, Quartey, Mengesha,
  Toups, Rickford, Jurafsky, and Goel}]{koenecke2020racial}
Allison Koenecke, Andrew Nam, Emily Lake, Joe Nudell, Minnie Quartey, Zion
  Mengesha, Connor Toups, John~R Rickford, Dan Jurafsky, and Sharad Goel. 2020.
\newblock Racial disparities in automated speech recognition.
\newblock \emph{Proceedings of the National Academy of Sciences},
  117(14):7684--7689.

\bibitem[{Krishna et~al.(2020)Krishna, Wieting, and
  Iyyer}]{krishna2020reformulating}
Kalpesh Krishna, John Wieting, and Mohit Iyyer. 2020.
\newblock \href {https://doi.org/10.18653/v1/2020.emnlp-main.55} {Reformulating
  unsupervised style transfer as paraphrase generation}.
\newblock In \emph{Proceedings of the 2020 Conference on Empirical Methods in
  Natural Language Processing (EMNLP)}, pages 737--762, Online. Association for
  Computational Linguistics.

\bibitem[{Labov(1972)}]{labov1972language}
William Labov. 1972.
\newblock \emph{Language in the inner city: Studies in the Black English
  vernacular}.
\newblock 3. University of Pennsylvania Press.

\bibitem[{Labov(1995)}]{labov1995case}
William Labov. 1995.
\newblock The case of the missing copula: The interpretation of zeros in
  african-american english.
\newblock \emph{Language}, 1:25--54.

\bibitem[{Labov et~al.(1998)}]{labov1998co}
William Labov et~al. 1998.
\newblock Co-existent systems in african-american vernacular english.
\newblock \emph{African-American English: Structure, History and Use}, pages
  110--153.

\bibitem[{Lample et~al.(2019)Lample, Subramanian, Smith, Denoyer, Ranzato, and
  Boureau}]{lample2018multiple}
Guillaume Lample, Sandeep Subramanian, Eric~Michael Smith, Ludovic Denoyer,
  Marc'Aurelio Ranzato, and Y{-}Lan Boureau. 2019.
\newblock \href {https://openreview.net/forum?id=H1g2NhC5KQ}
  {Multiple-attribute text rewriting}.
\newblock In \emph{7th International Conference on Learning Representations,
  {ICLR} 2019, New Orleans, LA, USA, May 6-9, 2019}. OpenReview.net.

\bibitem[{Lange(2012)}]{lange2012syntax}
Claudia Lange. 2012.
\newblock \emph{The syntax of spoken Indian English}, volume~45.
\newblock John Benjamins Publishing.

\bibitem[{Li and Liang(2021)}]{li2021prefixtuning}
Xiang~Lisa Li and Percy Liang. 2021.
\newblock \href {https://doi.org/10.18653/v1/2021.acl-long.353} {Prefix-tuning:
  Optimizing continuous prompts for generation}.
\newblock In \emph{Proceedings of the 59th Annual Meeting of the Association
  for Computational Linguistics and the 11th International Joint Conference on
  Natural Language Processing (Volume 1: Long Papers)}, pages 4582--4597,
  Online. Association for Computational Linguistics.

\bibitem[{Lippi-Green(1997)}]{lippi1997we}
Rosina Lippi-Green. 1997.
\newblock What we talk about when we talk about ebonics: Why definitions
  matter.
\newblock \emph{The Black Scholar}, 27(2):7--11.

\bibitem[{Liu et~al.(2019)Liu, Ott, Goyal, Du, Joshi, Chen, Levy, Lewis,
  Zettlemoyer, and Stoyanov}]{liu2019roberta}
Yinhan Liu, Myle Ott, Naman Goyal, Jingfei Du, Mandar Joshi, Danqi Chen, Omer
  Levy, Mike Lewis, Luke Zettlemoyer, and Veselin Stoyanov. 2019.
\newblock \href {http://arxiv.org/abs/1907.11692} {Roberta: A robustly
  optimized bert pretraining approach}.

\bibitem[{Martin et~al.(1998)Martin, Wolfram et~al.}]{martin1998sentence}
Stefan Martin, Walt Wolfram, et~al. 1998.
\newblock The sentence in african-american vernacular english.
\newblock \emph{African American English: structure, history, and use}, pages
  11--36.

\bibitem[{McCoy et~al.(2019)McCoy, Pavlick, and Linzen}]{mccoy-etal-2019-right}
Tom McCoy, Ellie Pavlick, and Tal Linzen. 2019.
\newblock \href {https://doi.org/10.18653/v1/P19-1334} {Right for the wrong
  reasons: Diagnosing syntactic heuristics in natural language inference}.
\newblock In \emph{Proceedings of the 57th Annual Meeting of the Association
  for Computational Linguistics}, pages 3428--3448, Florence, Italy.
  Association for Computational Linguistics.

\bibitem[{Mikolov et~al.(2013)Mikolov, Sutskever, Chen, Corrado, and
  Dean}]{mikolov2013distributed}
Tom{\'{a}}s Mikolov, Ilya Sutskever, Kai Chen, Gregory~S. Corrado, and Jeffrey
  Dean. 2013.
\newblock \href
  {https://proceedings.neurips.cc/paper/2013/hash/9aa42b31882ec039965f3c4923ce901b-Abstract.html}
  {Distributed representations of words and phrases and their
  compositionality}.
\newblock In \emph{Advances in Neural Information Processing Systems 26: 27th
  Annual Conference on Neural Information Processing Systems 2013. Proceedings
  of a meeting held December 5-8, 2013, Lake Tahoe, Nevada, United States},
  pages 3111--3119.

\bibitem[{Nguyen et~al.(2016)Nguyen, Do{\u{g}}ru{\"o}z, Ros{\'e}, and
  De~Jong}]{nguyen2016computational}
Dong Nguyen, A~Seza Do{\u{g}}ru{\"o}z, Carolyn~P Ros{\'e}, and Franciska
  De~Jong. 2016.
\newblock Computational sociolinguistics: A survey.
\newblock \emph{Computational linguistics}, 42(3):537--593.

\bibitem[{Nie et~al.(2020)Nie, Williams, Dinan, Bansal, Weston, and
  Kiela}]{nie-etal-2020-adversarial}
Yixin Nie, Adina Williams, Emily Dinan, Mohit Bansal, Jason Weston, and Douwe
  Kiela. 2020.
\newblock \href {https://doi.org/10.18653/v1/2020.acl-main.441} {Adversarial
  {NLI}: A new benchmark for natural language understanding}.
\newblock In \emph{Proceedings of the 58th Annual Meeting of the Association
  for Computational Linguistics}, pages 4885--4901, Online. Association for
  Computational Linguistics.

\bibitem[{Rajpurkar et~al.(2016)Rajpurkar, Zhang, Lopyrev, and
  Liang}]{rajpurkar2016squad}
Pranav Rajpurkar, Jian Zhang, Konstantin Lopyrev, and Percy Liang. 2016.
\newblock \href {https://doi.org/10.18653/v1/D16-1264} {{SQ}u{AD}: 100,000+
  questions for machine comprehension of text}.
\newblock In \emph{Proceedings of the 2016 Conference on Empirical Methods in
  Natural Language Processing}, pages 2383--2392, Austin, Texas. Association
  for Computational Linguistics.

\bibitem[{Reddy et~al.(2019)Reddy, Chen, and Manning}]{reddy2019coqa}
Siva Reddy, Danqi Chen, and Christopher~D. Manning. 2019.
\newblock \href {https://doi.org/10.1162/tacl_a_00266} {{C}o{QA}: A
  conversational question answering challenge}.
\newblock \emph{Transactions of the Association for Computational Linguistics},
  7:249--266.

\bibitem[{Reyes(2005)}]{reyes2005appropriation}
Angela Reyes. 2005.
\newblock Appropriation of african american slang by asian american youth 1.
\newblock \emph{Journal of Sociolinguistics}, 9(4):509--532.

\bibitem[{Ribeiro et~al.(2018)Ribeiro, Singh, and
  Guestrin}]{ribeiro2018semantically}
Marco~Tulio Ribeiro, Sameer Singh, and Carlos Guestrin. 2018.
\newblock \href {https://doi.org/10.18653/v1/P18-1079} {Semantically equivalent
  adversarial rules for debugging {NLP} models}.
\newblock In \emph{Proceedings of the 56th Annual Meeting of the Association
  for Computational Linguistics (Volume 1: Long Papers)}, pages 856--865,
  Melbourne, Australia. Association for Computational Linguistics.

\bibitem[{Rickford(2002)}]{rickford2002linguists}
John~R Rickford. 2002.
\newblock How linguists approach the study of language and dialect.
\newblock \emph{Ms. January}.

\bibitem[{Rickford(2016)}]{rickford2016raciolinguistics}
John~R Rickford. 2016.
\newblock \emph{Raciolinguistics: How language shapes our ideas about race}.
\newblock Oxford University Press.

\bibitem[{Rickford and King(2016)}]{rickford2016language}
John~R Rickford and Sharese King. 2016.
\newblock Language and linguistics on trial: Hearing rachel jeantel (and other
  vernacular speakers) in the courtroom and beyond.
\newblock \emph{Language}, 92(4):948--988.

\bibitem[{Rios(2020)}]{rios2020fuzze}
Anthony Rios. 2020.
\newblock Fuzze: Fuzzy fairness evaluation of offensive language classifiers on
  african-american english.
\newblock In \emph{Proceedings of the AAAI Conference on Artificial
  Intelligence}, volume~34, pages 881--889.

\bibitem[{Schuler and Namioka(1993)}]{schuler1993participatory}
Douglas Schuler and Aki Namioka. 1993.
\newblock \emph{Participatory design: Principles and practices}.
\newblock CRC Press.

\bibitem[{Shoemark et~al.(2018)Shoemark, Kirby, and
  Goldwater}]{shoemark2018inducing}
Philippa Shoemark, James Kirby, and Sharon Goldwater. 2018.
\newblock \href {https://doi.org/10.18653/v1/W18-6101} {Inducing a lexicon of
  sociolinguistic variables from code-mixed text}.
\newblock In \emph{Proceedings of the 2018 {EMNLP} Workshop W-{NUT}: The 4th
  Workshop on Noisy User-generated Text}, pages 1--6, Brussels, Belgium.
  Association for Computational Linguistics.

\bibitem[{Sidnell(2002)}]{sidnell2002african}
Jack Sidnell. 2002.
\newblock African american vernacular english (aave) grammar. 1.7.
\newblock \emph{Retrieved April}, 19(2009):16.

\bibitem[{Spears et~al.(1998)}]{spears1998african}
Arthur~K Spears et~al. 1998.
\newblock African-american language use: Ideology and so-called obscenity.
\newblock \emph{African-American English: Structure, history, and use}, pages
  226--250.

\bibitem[{Stewart(2014)}]{stewart2014now}
Ian Stewart. 2014.
\newblock \href {https://doi.org/10.3115/v1/E14-3004} {Now we stronger than
  ever: {A}frican-{A}merican {E}nglish syntax in {T}witter}.
\newblock In \emph{Proceedings of the Student Research Workshop at the 14th
  Conference of the {E}uropean Chapter of the Association for Computational
  Linguistics}, pages 31--37, Gothenburg, Sweden. Association for Computational
  Linguistics.

\bibitem[{Tan et~al.(2020)Tan, Joty, Kan, and Socher}]{tan2020s}
Samson Tan, Shafiq Joty, Min-Yen Kan, and Richard Socher. 2020.
\newblock \href {https://doi.org/10.18653/v1/2020.acl-main.263} {It{'}s
  morphin{'} time! {C}ombating linguistic discrimination with inflectional
  perturbations}.
\newblock In \emph{Proceedings of the 58th Annual Meeting of the Association
  for Computational Linguistics}, pages 2920--2935, Online. Association for
  Computational Linguistics.

\bibitem[{Thompson(2016)}]{thompson2016morpho}
Deanna Thompson. 2016.
\newblock The morpho-syntax of aspectual stay in aave.

\bibitem[{Trotta and Blyahher(2011)}]{trotta2011game}
Joe Trotta and Oleg Blyahher. 2011.
\newblock Game done changed: A look at selected aave features in the tv series
  the wire.
\newblock \emph{Moderna spr{\aa}k}, 105(1):15--42.

\bibitem[{Wang et~al.(2019)Wang, Singh, Michael, Hill, Levy, and
  Bowman}]{wang2018glue}
Alex Wang, Amanpreet Singh, Julian Michael, Felix Hill, Omer Levy, and
  Samuel~R. Bowman. 2019.
\newblock \href {https://openreview.net/forum?id=rJ4km2R5t7} {{GLUE:} {A}
  multi-task benchmark and analysis platform for natural language
  understanding}.
\newblock In \emph{7th International Conference on Learning Representations,
  {ICLR} 2019, New Orleans, LA, USA, May 6-9, 2019}. OpenReview.net.

\bibitem[{Warstadt et~al.(2019)Warstadt, Singh, and
  Bowman}]{warstadt2018neural}
Alex Warstadt, Amanpreet Singh, and Samuel~R. Bowman. 2019.
\newblock \href {https://doi.org/10.1162/tacl_a_00290} {Neural network
  acceptability judgments}.
\newblock \emph{Transactions of the Association for Computational Linguistics},
  7:625--641.

\bibitem[{Wee(2008)}]{wee2008singapore}
Lionel Wee. 2008.
\newblock Singapore english: morphology and syntax.
\newblock In \emph{A handbook of varieties of English}, pages 2250--2264. De
  Gruyter Mouton.

\bibitem[{Wolfram and Schilling(2015)}]{wolfram2015american}
Walt Wolfram and Natalie Schilling. 2015.
\newblock \emph{American English: dialects and variation}.
\newblock John Wiley \& Sons.

\bibitem[{Zalmout et~al.(2018)Zalmout, Erdmann, and Habash}]{zalmout2018noise}
Nasser Zalmout, Alexander Erdmann, and Nizar Habash. 2018.
\newblock \href {https://doi.org/10.18653/v1/N18-1087} {Noise-robust
  morphological disambiguation for dialectal {A}rabic}.
\newblock In \emph{Proceedings of the 2018 Conference of the North {A}merican
  Chapter of the Association for Computational Linguistics: Human Language
  Technologies, Volume 1 (Long Papers)}, pages 953--964, New Orleans,
  Louisiana. Association for Computational Linguistics.

\bibitem[{Zampieri et~al.(2014)Zampieri, Tan, Ljube{\v{s}}i{\'c}, and
  Tiedemann}]{zampieri2014report}
Marcos Zampieri, Liling Tan, Nikola Ljube{\v{s}}i{\'c}, and J{\"o}rg Tiedemann.
  2014.
\newblock \href {https://doi.org/10.3115/v1/W14-5307} {A report on the {DSL}
  shared task 2014}.
\newblock In \emph{Proceedings of the First Workshop on Applying {NLP} Tools to
  Similar Languages, Varieties and Dialects}, pages 58--67, Dublin, Ireland.
  Association for Computational Linguistics and Dublin City University.

\bibitem[{Zhang et~al.(2019)Zhang, Baldridge, and He}]{zhang-etal-2019-paws}
Yuan Zhang, Jason Baldridge, and Luheng He. 2019.
\newblock \href {https://doi.org/10.18653/v1/N19-1131} {{PAWS}: Paraphrase
  adversaries from word scrambling}.
\newblock In \emph{Proceedings of the 2019 Conference of the North {A}merican
  Chapter of the Association for Computational Linguistics: Human Language
  Technologies, Volume 1 (Long and Short Papers)}, pages 1298--1308,
  Minneapolis, Minnesota. Association for Computational Linguistics.

\end{thebibliography}

\appendix

\section{Details on the Transformation Rules}
\subsection{Morphosyntactic translation} 
\label{appdx:morphosyntax}
We conduct morphosyntactic translation as the first step in the pipeline because our methods are based on grammatical rules that are determined by an SAE dependency parse. Here, we provide further details for our methods. We rely on \texttt{spaCy} to dependency parse the GLUE text at the sentence level before proceeding.

\paragraph{Inflection.} In AAVE, speakers do not inflect simple present or past tense verbs differently for number or person \cite{green2002african}. This means the SAE sentence ``\textit{She studies linguistics}'' would be rendered in AAVE as ``\textit{She \textbf{study} linguistics}.'' We identify all regular present verbs by their \texttt{VBZ} or \texttt{VBP} part of speech tag, and regular past verbs by their \texttt{VBD} part of speech tag. Then we inflect these verbs to standard first-person \texttt{VBP} or \texttt{VBD} respectively, using the \texttt{pyinflect} library.

\paragraph{Auxiliaries.} In AAVE, auxiliaries with negated heads can be replaced by \textit{ain't} \cite{green2002african}, and we make this conversion first. Copula deletion and optional auxiliary dropping are also grammatical in AAVE \cite{stewart2014now,green2002african,labov1972language,wolfram2015american}. This means the SAE sentence ``\textit{We are better than before}'' would be rendered in AAVE without the copula as ``\textit{We better than before}.'' The question ``\textit{Did you see that?}'' could be rendered without the auxiliary verb as ``\textit{You see that?}'' Similarly, the phrase ``\textit{I have seen him}'' could go without the auxiliary, as in ``\textit{I seen him}.'' 

We treat the dropped copula as a separate case. Since it only applies to the present tense \textit{is} and \textit{are}, we search for these tokens and check that the environment is one where \textit{contraction} would be allowed in SAE. We ensure the copula is not negated and that it has an object dependant; that is neither a clausal complement nor the head of a clausal complement. We have confirmed that these decisions all account for the fact that copula deletion is disallowed in non-finite contexts, imperatives, ellipsis, inversion environments, or complement and subject extraction environments \cite{bender2000syntactic,labov1995case}.

To account for other auxiliary dropping, we drop tokens with the \texttt{AUX} part of speech tag. We do not drop modals (tag \texttt{MD}), the future tense marker \textit{will}, or any token whose head is a copula or an open clausal complement (\texttt{xcomp}). 

\paragraph{Existential \textit{dey}/\textit{it}.} AAVE speakers can indicate something exists by using what is known as an \textit{it} or \textit{dey} existential construction \cite{green2002african}. The existential construction in ``\textit{It's some milk in the fridge}'' is used to mean ``\textit{There is some milk in the fridge}.'' We make this transformation by searching the text for expletive or pleonastic nominals (\texttt{expl} dependencies) and substituting these tokens with either \textit{it} or \textit{dey} with equal probability.

\paragraph{Negative concord.} This phenomenon, also called \textit{multiple negation} or \textit{pleonastic negation}, is ``the use of two negative morphemes to communicate a single negation,'' a widely-known feature of AAVE \cite{martin1998sentence}. For example, the SAE sentence ``\textit{He doesn't have a camera}'' would become ``\textit{He don't have no camera}.'' To capture this transformation, we search the text for \texttt{neg} dependents of verbal heads. Then we negate the object dependents\footnote{'dobj', 'iobj', 'obj', 'pobj', 'obl', 'attr'} of that verbal head. In these constructions, the negation can only be marked on auxiliaries and indefinite nouns \cite{green2002african}, but not definite nouns. We check for indefiniteness by ensuring that the object itself has only indefinite determiner children (a/an), and that the object is not a proper noun (\texttt{NNP}), nor a personal pronoun (\texttt{PRP}), nor is it an adjective modifier (\texttt{amod}). We also ensure that the object is not already a Negative Polarity Item (e.g. \textit{nobody}, \textit{nothing}).

\paragraph{Negative inversion.} This AAVE feature is superficially similar to negative concord. Both an auxiliary and an indefinite noun phrase are negated at the beginning of a sentence or clause \cite{green2002african,martin1998sentence}. For example, the SAE assertion that ``\textit{no suffering lasts forever}'' would be rendered in AAVE as ``\textit{don't no suffering last forever.}'' Since there was no auxiliary already present to front and negate, the syntax required obligatory \textit{do} support. When the statement contains an auxiliary, the auxiliary verb will be fronted and negated instead, as in the transformation from ``\textit{Nobody can hear you}'' to ``\textit{Can't nobody hear you.}'' We operationalize these rules using the dependency parse. Specifically, we identify the span of the given clause by traveling up the dependency tree until we hit a ROOT, conjunction (\texttt{conj}), or complement dependency; then we use tree traversal from that origin to find the smallest index in the clause. In this way, we confirm that the negation is clause-initial.

\paragraph{Relative clause structures.} AAVE speakers most frequently use the complementizer \textit{that} to introduce relative clauses, rather than using Wh-pronouns like \textit{who, where, when} \cite{martin1998sentence}. There is also a grammatical option to drop the complementizer altogether. For example, ``\textit{There are a whole lot of people who don't want to go to hell}'' could become in AAVE, ``\textit{It's a whole lot of people don' wanna go to hell}'' \cite{green2002african}. In our transformation, we simply drop all lemmas \textit{who} and \textit{that}
where the head is a relative clause modifier (\texttt{relcl}).

\paragraph{Null genitives.} AAVE allows a null genitive marking \cite{stewart2014now,wolfram2015american}. For example, ``\textit{Rolanda's bed isn't made}'' can be rendered ``\textit{Rolanda bed don't be made up}'' \cite{green2002african}. To capture this pattern, we simply drop any possessive endings (\texttt{POS}) from the text.

\paragraph{Completive \textit{done} and remote time \textit{been}.} The phrase ``\textit{I had written it.}'' can be rendered in AAVE as ``\textit{I done wrote it}'' using the completive verbal marker d\textschwa{}n. The phrase ``\textit{He ate a long time ago}'' can be rendered as ``\textit{He been ate}'' using the remote time BIN \cite{green2002african}. To operationalize this construction, we search for simple past verbs (\texttt{VBD}) with temporal noun phrase adverbial modifier children (\texttt{npadvmod}), like the \textit{yesterday} in \textit{I ate yesterday.} Then appended either \textit{done} or \textit{been} preverbally, each with equal probability. We also consider past participle verbs (\texttt{VBN}), and we replace the \textit{have} auxiliaries with \textit{done}/\textit{been}.

\paragraph{\textit{Ass} constructions.} These constructions may be mis-classified as obscenity, but they serve a distinct and consistent role in AAVE grammar \cite{spears1998african}. One common form called the \textit{ass camouflage construction} \cite{collins2008aae} can be seen in the phrase ``\textit{I divorced his ass}.'' Here it behaves as a metonymic pseudo-pronoun \cite{spears1998african}. Similarly, the form can appear reflexively, as in ``\textit{Get yo'ass inside}.'' \textit{Ass} constructions can also serve as discourse-level expressive markers or intensifiers, as in the compound ``\textit{We was at some random-ass bar}.'' To operationalize the former, we substitute the appropriate \textit{ass} construction for any personal pronoun (\texttt{PRP}) that was the object of a verb. To operationalize the latter, we transform adjective modifiers (\texttt{amod}). Not all adjectives can participate in this construction, however. That is why we consider only \textit{gradable} adjectives \cite{kennedy2007vagueness}, or adjectives that accept comparative and superlative modifiers and morphology. For example, \textit{cold} can become \textit{colder, very cold, coldest}, so a \textit{cold-ass day} is an acceptable phrase in AAVE. Non-gradable or absolute adjectives like \textit{finished} and \textit{American} cannot participate; it is not acceptable to say \textit{this finished-ass project} or \textit{that American-ass woman} in AAVE.

\paragraph{Future \textit{gonna} and immediate future \textit{finna}.} In AAVE, the future tense is marked by \textit{gon} or \textit{gonna} instead of \textit{will}, as in ``\textit{You gon understand}'' \cite{green2002african,sidnell2002african}. In the first person, this becomes \textit{I'ma}. In the immediate future, speakers can use \textit{finna} (or variants \textit{fixina, fixna and fitna}), as in ``\textit{I'm finna leave.}'' Although they are morphosyntactic, we treat these cases with simple lexical substitution.

\paragraph{Have / got.} In the casual speech of AAVE and other dialects, both the modal and the verb form of \textit{have} can be replaced by \textit{got} \cite{trotta2011game}. \textit{Have to} can become \textit{got to} or \textit{gotta}, and similar for the verb of possession. We simply convert the present-tense \textit{have} and \textit{has} to \textit{got} and ensure that the verb has an object.

\subsection{Lexical and orthographic translation}
\label{appdx:lexicon}

The seed list from \citet{shoemark2018inducing} contained the (1) \textit{the/tha}, (2) \textit{with/wit}, (3) \textit{getting/gettin}, (4) \textit{just/jus}, (5) \textit{and/nd}, (6) \textit{making/makin}, (7) \textit{when/wen}, (8) \textit{looking/lookin}, (9) \textit{something/somethin}, (10) \textit{going/goin}.



\section{Lightweight Training}
\label{appdx:prefix_tuning}
Directly training new models for every language variety is expensive in both compute time and storage space. This motivates a lightweight fine-tuning strategy inspired by the state-of-the-art prefix-tuning method \cite{li2021prefixtuning}. Specifically, we freeze the models trained on SAE. Then, for each dialect $d$, we fine-tune a transformation matrix $M_d$. 
When training the dialect-specific model, we append $M_d$ to the embeddings $e_i^d$ of each input sequence $x_i^d$. 
The matrix $M_d$ is the only parameter that needs to be trained and stored besides the base model. 
During inference on dialect $d$, we can directly fetch base SAE model 
and the corresponding transformation matrix $M_d$ to form the dialect-specific model 
and make predictions. Besides efficient domain adaptation, one additional advantage may be improved out-of-domain generalization \cite{li2021prefixtuning}. Following \citet{li2021prefixtuning}, we used a batch size of 16. The prefix length was set to 50; the maximum learning rate was $5e-4$; the maximum number of training epochs was $5$.

Results are given in Table~\ref{tab:lightweight_performance}. The second row for each task is labeled Prefix Tuning, and it gives the results of our lightweight fine-tuning approach. Prefix tuning demonstrates reasonable performance, but, with the exception of SST-2 sentiment analysis, Prefix Tuning fails to match the performance of full AAVE (\data{}) training. Thus there is still a need for more effective and efficient domain adaption methods for dialects like AAVE.

\begin{table}[h!]
\resizebox{\columnwidth}{!}{%
\def\arraystretch{1.15}
\begin{tabular}{clcc}\toprule
 &&\multicolumn{2}{c}{Synthetic Testing}
 \\ 
 & Training & SAE & AAVE \\ \midrule
 
 \textbf{CoLA} & AAVE (\data) & \textbf{56.2} & \textbf{55.8}\\
 & Prefix Tuning & 17.0 & 17.1 \\ 
 \midrule
 
 \textbf{MNLI} & AAVE (\data) & \textbf{83.1} & \textbf{83.5} \\
 & Prefix Tuning & 82.1 & 81.5\\ 
 \midrule
 
 \textbf{QNLI} & AAVE (\data) & \textbf{92.5} & \textbf{91.8} \\
 & Prefix Tuning & 86.7& 86.0 \\ 
 \midrule
 
 \textbf{RTE} & AAVE (\data) & \textbf{67.1} & \textbf{67.2} \\
 & Prefix Tuning & 54.5 & 54.1\\ \midrule
 
 \textbf{SST-2} & AAVE (\data)  & 94.0	 & 93.0\\
 & Prefix Tuning & \textbf{94.6}	& \textbf{93.1} \\ \midrule
 
 \textbf{STS-B} & AAVE (\data) & \textbf{88.8}	& \textbf{88.3}\\
 & Prefix Tuning & 27.8	 & 25.1 \\ \hline
 
 \textbf{QQP} & AAVE (\data) & \textbf{90.3} & \textbf{89.8}\\
 & Prefix Tuning & 88.7 & 88.0 \\ \hline
\bottomrule 
\end{tabular}
}
\caption{\textbf{Lightweight tuning results} for six tasks (Matthew's Corr. for CoLA; Pearson-Spearman Corr. for STS-B; Accuracy for all others). Prefix Tuning fails to match the performance of full AAVE (\data{}) training
}
\label{tab:lightweight_performance}
\end{table}

\section{Detailed Examples from the Perturbation Analysis}
\label{appdx:perturbation_analysis}

For each transformation type, we provide an example of each error category in Tables \ref{tab:perturbation_error_examples_1} and \ref{tab:perturbation_error_examples_2} when applicable. Here, we will briefly discuss our observations. For \textit{aux-dropping}, the most common error is to confuse neutral relationships for contradictions (\textbf{n$\rightarrow$c}). The model may fail to link the subject with the predicate of the \textsc{Hypothesis} without the overt copula. We also notice an entailment relation mistaken for a contradiction when the loss of the auxiliary verb renders \textit{mine mutts} syntactically ambiguous. Its position in the sentence suggests a Noun Phrase where the possessive pronoun \textit{mine} is used in place of the possessive adjective \textit{my}.\footnote{This usage can be seen in sources such as the KJV Bible and in older hymns like the 1862 \textit{Mine Eyes Have Seen the Glory}.} For \textit{completive done} and \textit{remote time been}, \textbf{n$\rightarrow$c} is the most common error due again, possibly due to a failure to link subject and predicate. However, the converse error \textbf{c$\rightarrow$n} may be triggered for similar reasons, as in \textit{The woman done never spoke before.} For both the \textit{inflection} rules and the \textit{lexical} changes, the most common error is to mistake an entailment relationship for a neutral one. This may be due to the fragmenting of common subsequences and an overall reduced lexical similarity between the \textsc{Premise} and the \textsc{Hypothesis}. Both lexical overlap and subsequence matching are well-known heuristics for NLI \cite{mccoy-etal-2019-right}. Finally, we recognize that some errors may arise from semantically ambiguous transformations. For example, in the \textbf{c$\rightarrow$n} Lexical error, the word \textit{right} was swapped for the alternative spelling \textit{rite}, which is misleading in the context of church, since \textit{rite} typically refers to a religious or ceremonial act. The transformation is not technically erroneous, but the setting renders it unfairly ambiguous.

\begin{table*}[t]
\resizebox{\textwidth}{!}{
\begin{tabular}{c|l|L{60mm}|L{110mm}}
\toprule
\textbf{Transf.} & \textbf{Error} & \textsc{Premise} & \textsc{Hypothesis}\\ \midrule
\multirow{13}{*}{\rotatebox[origin=c]{90}{Auxiliaries}} & \textbf{c}$\rightarrow$\textbf{n} & Energy-related activities are the primary source of U.S. man-made greenhouse gas emissions. & {\textit{Producing cars is the main source of US greenhouse gas emissions.} \newline $\hookrightarrow$ Producing cars the main source of US greenhouse gas emissions}.\\ \cmidrule{2-4}
& \textbf{c}$\rightarrow$\textbf{e} & Search out the House of Dionysos and the House of the Trident with their simple floor patterns, and the House of Dolphins and the House of Masks for more elaborate examples, including Dionysos riding a panther, on the floor of the House of Masks. & {\textit{The floor patterns of the House of the Trident are very intricate.} \newline $\hookrightarrow$ The floor patterns of the House of the Trident very intricate.} \\ \cmidrule{2-4}
& \textbf{n}$\rightarrow$\textbf{c} & To the west of the city at Hillend is Midlothian Ski Centre, the longest artificial ski slope in Europe. & {\textit{The Midlothian Ski Centre is the only artificial ski slope in Scotland.} \newline $\hookrightarrow$ The Midlothian Ski Centre the only artificial ski slope in Scotland.}\\ \cmidrule{2-4}
& \textbf{n}$\rightarrow$\textbf{e} & Mykonos has had a head start as far as diving is concerned because it was never banned here (after all, there are no ancient sites to protect). & {\textit{Protection of ancient sites is the reason for diving bans in other places.} \newline $\hookrightarrow$ Protection of ancient sites the reason for diving bans in other places.} \\ \cmidrule{2-4}
& \textbf{e}$\rightarrow$\textbf{c} & oh yeah all all mine are uh purebreds so i keep them in & {\textit{none of mine are mutts} \newline $\hookrightarrow$ none of mine mutts} \\ \cmidrule{2-4}
& \textbf{e}$\rightarrow$\textbf{n} & This particular instance of it stinks. & {\textit{It is a terrible situation.} \newline $\hookrightarrow$ It a terrible situation.} \\ \midrule
\multirow{11}{*}{\rotatebox[origin=c]{90}{Been / done}} &\textbf{c}$\rightarrow$\textbf{n} & She had spoken with no trace of foreign accent. & {\textit{The woman had never spoken before.} \newline $\hookrightarrow$ The woman done never spoken before.} \\ \cmidrule{2-4}
&\textbf{n}$\rightarrow$\textbf{c} & For more than 26 centuries it has witnessed countless declines, falls, and rebirths, and today continues to resist the assaults of brutal modernity in its time-locked, color-rich historical center. & {\textit{Modernity has made no progress in the historical center.} \newline $\hookrightarrow$ Modernity done made no progress in the historical center.}\\ \cmidrule{2-4}
&\textbf{n}$\rightarrow$\textbf{e} & (And yes, he has said a few things that can, with some effort, be construed as support for supply-side economics.) & {\textit{He has begun working on construing the things as support for supply-side economics.} \newline $\hookrightarrow$ He done begun working on construing the things as support for supply-side economics.} \\ \cmidrule{2-4}
&\textbf{e}$\rightarrow$\textbf{c} & Detroit Pistons they're not as good as they were last year & {\textit{Detroit Pistons played better last year} \newline $\hookrightarrow$ Detroit Pistons done played better last year} \\ \cmidrule{2-4}
&\textbf{e}$\rightarrow$\textbf{n} & I don't know what I would have done without Legal Services, said James. & {\textit{James said Legal Services helped him a lot.} \newline $\hookrightarrow$ James said Legal Services been helped him a lot.}\\ \midrule
\multirow{12}{*}{\rotatebox[origin=c]{90}{Inflection}} &\textbf{c}$\rightarrow$\textbf{n} &  Once or twice, but they seem more show than battle, said Adrin. & {\textit{Adrin said they were amazing warriors.} \newline $\hookrightarrow$ Adrin said they was amazing warriors.} \\ \cmidrule{2-4}
&\textbf{c}$\rightarrow$\textbf{e} &   The story of the technology business gets spiced up because the reality is so bland.& {\textit{Reality is so bland that the garbage business gets spiced up.} \newline $\hookrightarrow$ Reality is so bland that the garbage business get spiced up.} \\ \cmidrule{2-4}
&\textbf{n}$\rightarrow$\textbf{c} & The air is warm. & {\textit{The arid air permeates the surrounding land.} \newline $\hookrightarrow$ The arid air permeate the surrounding land.} \\ \cmidrule{2-4}
&\textbf{n}$\rightarrow$\textbf{e} & Long ago--or away, or whatever--there was a world called Thar?? and another called Erath. & {\textit{Thar and Erath were not the only worlds in existence then.} \newline $\hookrightarrow$ Thar and Erath was not the only worlds in existence then.} \\ \cmidrule{2-4}
&\textbf{e}$\rightarrow$\textbf{c} & The disputes among nobles were not the first concern of ordinary French citizens. & {\textit{Ordinary French citizens were not concerned with the disputes among nobles.} \newline $\hookrightarrow$ Ordinary French citizens was not concerned with the disputes among nobles.} \\ \cmidrule{2-4}
&\textbf{e}$\rightarrow$\textbf{n} & Perched on a steep slope, high in the Galilean hills, Safed (known also as Tzfat, Tsfat, Sefat, and Zefat) is a delightful village-town of some 22,000 people. & {\textit{Safed is a village that goes by numerous other names.} \newline $\hookrightarrow$ Safed is a village that go by numerous other names.}
\\ \bottomrule
\end{tabular}}
\caption{\small{\textbf{Example perturbation errors} for \textit{aux-dropping, been/done,} and \textit{inflection} transformations. The \textsc{Hypothesis} was transformed from the original \textit{SAE} $\hookrightarrow$ Synthetic AAVE.}
} \label{tab:perturbation_error_examples_1}
\end{table*}

\begin{table*}[t]
\resizebox{\textwidth}{!}{
\begin{tabular}{c|l|L{100mm}|L{110mm}}
\toprule
\textbf{Transf.} & \textbf{Error} & \textsc{Premise} & \textsc{Hypothesis}\\ \midrule
\multirow{12}{*}{\rotatebox[origin=c]{90}{Lexical}} &\textbf{c}$\rightarrow$\textbf{n} & Oh, sorry, wrong church. & {\textit{It was the right church.} \newline $\hookrightarrow$ It was the rite church} \\ \cmidrule{2-4}
&\textbf{c}$\rightarrow$\textbf{e} &  The story of the technology business gets spiced up because the reality is so bland. & {\textit{Reality is so bland that the garbage business gets spiced up.} \newline $\hookrightarrow$ Reality is so bland that the garbage bizness gets spiced up.} \\ \cmidrule{2-4}
&\textbf{n}$\rightarrow$\textbf{c} & And who should decide? & {\textit{No one is willing to make the decision.} \newline $\hookrightarrow$ No one is willin to make the decision.} \\ \cmidrule{2-4}
&\textbf{n}$\rightarrow$\textbf{e} &  At the eastern end of Back Lane and turning right, Nicholas Street becomes Patrick Street, and in St. Patrick's Close is St. Patrick's Cathedral. & {\textit{Back Lane and Nicholas Street are longer than Patrick Street.} \newline $\hookrightarrow$ Bacc Lane and Nicholas Street r longer than Patrick Street.} \\ \cmidrule{2-4}
&\textbf{e}$\rightarrow$\textbf{c} & Even analysts who had argued for loosening the old standards, by which the market was clearly overvalued, now think it has maxed out for a while. & {\textit{Some analysts wanted to make the old standards easier.} \newline $\hookrightarrow$ Sum analysts wanted to make the old standards easier.} \\ \cmidrule{2-4}
&\textbf{e}$\rightarrow$\textbf{n} & 8 million in relief in the form of emergency housing. & {\textit{Emergency housing relief totaled 8 million dollars.} \newline $\hookrightarrow$ Emergency housing relief totaled 8 million dollas.} \\ \midrule
\multirow{19}{*}{\rotatebox[origin=c]{90}{Negative Concord}} &\textbf{c}$\rightarrow$\textbf{n} & Agencies may perform the analyses required by sections 603 and 604 in conjunction with or as part of any other agenda or analysis required by other law if such other analysis satisfies the provisions of these sections. & {\textit{The agency is free to decide not to perform the analyses covered in section 603.} \newline $\hookrightarrow$ The agency is free to decide not to perform no analyses covered in section 603.}\\ \cmidrule{2-4}
&\textbf{c}$\rightarrow$\textbf{e} & and clean up is is uh is a joy uh a little soap and water and air dry them and you don't have to worry about that & {\textit{You don't need any soap for the clean up.} \newline $\hookrightarrow$ You don't need no soap for the clean up.}  \\ \cmidrule{2-4}
&\textbf{n}$\rightarrow$\textbf{c} & Each state is different, and in some states, intra-state regions differ significantly as well. & {\textit{You can go from one area of a state to another and not see a resemblance.} \newline $\hookrightarrow$ You can go from one area of a state to another and not see no resemblance} \\ \cmidrule{2-4}
&\textbf{n}$\rightarrow$\textbf{e} & Bars with views and live music include Sky Lounge in the Sheraton Hotel and Towers, Tsim Sha Tsui; and Cyrano in the Island Shangri-La in Pacific Place. & {\textit{It is not far to go to find good drink, gorgeous views, and live music.} \newline $\hookrightarrow$ It ain't far to go to find good drink, gorgeous views, and live music.} \\ \cmidrule{2-4}
&\textbf{e}$\rightarrow$\textbf{c} & Regulators may not be totally supportive of a more comprehensive business model because they are concerned that the information would be based on a lot of judgment and, therefore, lack of precision, which could make enforcement of reporting standards difficult. & {\textit{Being totally supportive of a more comprehensive business model is not something regulators may do.} \newline $\hookrightarrow$ Being totally supportive of a more comprehensive business model ain't something regulators may do.} \\ \cmidrule{2-4}
&\textbf{e}$\rightarrow$\textbf{n} &  uh well no i just know i know several single mothers who absolutely can't afford it they have to go with the a single uh what i mean a babysitter more more or less & {\textit{They simply don't have the money to put into that sort of thing.} \newline $\hookrightarrow$ They simply don't have no money to put into that sort of thing.}\\ \midrule
\multirow{15}{*}{\rotatebox[origin=c]{90}{Relative Clause Structures}} &\textbf{c}$\rightarrow$\textbf{n} &He unleashed a 16-day reign of terror that left 300 Madeirans dead, stocks of sugar destroyed, and the island plundered.  & {\textit{He unleashed a large debate over the 16-day reign that ended in a peaceful protest.} \newline $\hookrightarrow$ He unleashed a large debate over the 16-day reign ended in a peaceful protest.} \\ \cmidrule{2-4}
&\textbf{c}$\rightarrow$\textbf{e} & Shoot only the ones that face us, Jon had told Adrin. & {\textit{Shoot the ones that face us, Adrin told Jon} \newline $\hookrightarrow$ Shoot the ones face us, Adrin told Jon} \\ \cmidrule{2-4}
&\textbf{n}$\rightarrow$\textbf{c} & The conspiracy-minded allege that the chains also leverage their influence to persuade the big publishers to produce more blockbusters at the expense of moderate-selling books. & {\textit{Most people who read a book, tend to watch a film adaptation of it.} \newline $\hookrightarrow$ Most people read a book, tend to watch a film adaptation of it.} \\ \cmidrule{2-4}
&\textbf{n}$\rightarrow$\textbf{e} & Wear a nicely ventilated hat and keep to the shade in the street. & {\textit{The street has plenty of shade for those who want it.} \newline $\hookrightarrow$ The street has plenty of shade for those want it.} \\ \cmidrule{2-4}
&\textbf{e}$\rightarrow$\textbf{c} & External Validity The extent to which a finding applies (or can be generalized) to persons, objects, settings, or times other than those that were the subject of study. & {\textit{External Validity gets its name from the fact that what's being studied are people, things, and individuals who are outside of the study.} \newline $\hookrightarrow$ External Validity gets its name from the fact that what's being studied are people, things, and individuals are outside of the study.} \\ \cmidrule{2-4}
&\textbf{e}$\rightarrow$\textbf{n} & Perched on a steep slope, high in the Galilean hills, Safed (known also as Tzfat, Tsfat, Sefat, and Zefat) is a delightful village-town of some 22,000 people. & {\textit{Safed is a village that goes by numerous other names.} \newline $\hookrightarrow$ Safed is a village goes by numerous other names.}
\\ \bottomrule
\end{tabular}}
\caption{\small{\textbf{Example perturbation errors} for \textit{lexical} transformations as well as \textit{negative concord} and \textit{wh-dropping} (\textit{relative clause structures}). The \textsc{Hypothesis} was transformed from the original \textit{SAE} $\hookrightarrow$ Synthetic AAVE.}
} \label{tab:perturbation_error_examples_2}
\end{table*}

\section{Detailed Error Analysis}
\label{appdx:detailed_error_analysis}
Here, we provide a more detailed error analysis for our transforamtion system, organize by transformation rules.

\begin{itemize}
    \item \textit{Broader issues.} GLUE contains examples from journalism and news, which tends to use a more academic or formal register. Some of the annotators were not accustomed to seeing language variation in the body copy of a newspaper and so they identified stylistic errors that may have been grammatically well-formed. On the other hand, GLUE also contains purposeful disfluencies, which harm the performance of the syntactic parsers in our pipeline.

    \item \textit{Existential dey/it.} Some annotators held that this feature should not be present in questions, so the example \textit{is there a place?} should not be swapped with \textit{is it a place?} 
    
    \item \textit{Auxiliaries.} Copula dropping is not allowed in questions or in cases of ellipsis like ``\textit{I like Bill's new wine, but Max's old \textbf{\O} even better.}'' Similarly, with long-range dependencies in appositive phrases, we should not drop that be-verb, as done here: ``\textit{Hyperthymesia, or hyperthymesic syndrome, \textbf{\O} a disorder}.''
    
    \item \textit{Been / done.} Errors appeared in this rule's handling of negation, like in ``\textit{This was a persistent problem which has not been solved}'' becoming ``\textit{This was a persistent problem which done not been solved.}''
    
    \item \textit{Have / got.} Some annotators found this transformation unacceptable in the subjunctive mood. For example, ``\textit{If the United States has a female president, ...} became ``\textit{If the United States got a female president.}''
    
    \item \textit{Inflection.} Some errors with inflection occur when the POS tagger erroneously marks a noun as a root verb or similar. Also, inflection rules should not apply to idioms like ``\textit{just goes to show...}''
    
    \item \textit{Negative concord.} Again, this rule should not apply to idiomatic or phrasal constituents. Negative concord also does not apply to finite nouns, including demonstratives, like in this error, swapping ``\textit{Couldn't nothing in Siegel's work explain this perception}'' for ``\textit{Couldn't nothing in Siegel's work explain \textbf{no} perception.}''
    
    \item \textit{Null genitive.} The possessive \textit{'s} should not be dropped when doing so would lead to syntactic ambiguity. For example, ``\textit{How I see someone's deleted Instagram account?}'' would become ambiguous in ``\textit{How I see someone deleted Instagram account?}'' A similar situation arises with the complex object NP in ``\textit{to grab the old lady at the end of my aisle\textbf{'s} walker.}''
\end{itemize}

Through iterative discussion with the DataWorkers throughout the participatory design process, we also gleaned the following insights. We recognize that it is possible for a linguistic transformation to alter the connotative or social meaning of a text without altering the denotative semantic meaning. In the following example, the phrase ``done cut her off'' is linguistically acceptable. Furthermore, the truth conditional meanings of (1) and (2) are equivalent. However, the social connotations differ.
\begin{quote}
    (1) \small{\textbf{GLUE}: Beth didn't get angry with Sally, who had cut her off, because she stopped and counted to ten.}\\\\
    (2) \small{\textbf{\data{}}: Beth ain't get angry with Sally, done cut her off, because she stopped and counted to ten.}
\end{quote}
One undergraduate annotator explained that \textit{``Done cut her off''} would be used if, for instance, the speaker was getting mad while explaining the details, and just threw that small piece of information in their speech.

Annotators also identified examples in which certain features could apply, but the rules had not yet implemented in our system. For example, the habitual \textit{be} \cite{stewart2014now,green2002african,wolfram2015american,labov1998co} would apply in the sentence ``\textit{Guatemala \textbf{be} accepting the Pet passport as proof of vaccination.}'' In order to capture these missing features and widen the diversity of our test sets, we build the Gold Standard Test sets in Section~\ref{subsec:gold_test_set}.

\section{Qualitative Comparison Between VALUE and Style Transfer}
\label{appdx:qualitative_evalue_strap}

Examples 1-5 show STRAP's creative phrase-level transformations, like converting ``absurdly suspicious'' to the phrase ``weird as hell.'' However, STRAP can distort the text by hallucinations like in Examples 6-7. In (8), STRAP removes the name of the subject \textit{Cochran}, a valuable detail. The neural style-transfer can also produce erratic behaviors as we see in 9 and 10. VALUE on the other hand appears to better capture AAVE morphosyntax. We see future \textit{gon} in 11, negative concord in 12, copula dropping in 13, uninflection in 14, and an ass intensifier in 15, none of which are represented in the STRAP output. These are the primary advantages of our approach: (1) integrity of underlying constructs, and (2) linguistically attested features that can be systematically analyzed.

\begin{table*}[t]
\resizebox{\textwidth}{!}{%
\begin{tabular}{lp{70mm}p{70mm}p{70mm}}
\toprule
& \textbf{GLUE} & \textbf{VALUE} &
\textbf{STRAP} \\ \midrule
1. & Really, Boris, you are absurdly suspicious. & Really, Boris, you absurdly suspicious. & you look \textbf{weird as hell} Boris \\ \hline
2. &In alliance with the Jesuits, he weeded out corrupt clerics and what he considered too soft Umiliati order of Catholic laymen. & In alliance with da Jesuits, he weeded out corrupt clerics and wht he considered too soft-ass Umiliati order of Catholic laymen. & Solo son and a corrupt priest quickly rid his self of corrupt members of the Umiliati community, which he considered \textbf{too lenient}. \\ \hline
3. &Jon did not expect better reactions from most of the town folk.& Jon did not expect betta reactions from most of the town folk. & \textbf{Most ppl} in the city didnt show any better reaction \\ \hline
4. &No, she replied with perfect truth. &No, she replied wid perfect truth.& \textbf{she was very accurate with her answer} \\ \hline
5. &I was bewildered, but I knew that it was no good asking him to explain. & I was bewildered, but I knew dat it was no gud asking his ass to explain. & I'm confused but I know \textbf{this can't help but make me more frustrated}\\ \hline
6. &my wife used to plant a few snow peas i don't really care for snow peas well she just plants a few for herself & my wife used to plant a few snow peas i don't really care fa snow no peas well she jus plant a few for herself &She use to put some peas \textbf{in the house} that \textbf{she don't really like peas}\\ \hline
7.&The real market for filters isn't Mom and Dad--it's Dilbert's boss.&The real market fa filters ain't moma and Dad--it's Dilbert boss.&\textbf{Dilbert not a momma or a dad} lmao\\ \hline
8.&Cochran pontificated windily for the camera.&Cochran pontificated windily for da camera.& \textbf{Pontificated} loud af in the camera\\ \hline
9.&yeah but then sometimes you know yeah um it just went up the first of this month i'm paying uh seventy nine dollars a week for now&yea but then sometimes u know yeah um it just been went up the first of this month i'm payin uh seventy nine dollars a week for now&\textbf{yea, yea, yea, yea, yea, yea, yea, yea, yea, yea}. You know it's like that sometimes.\\ \hline
10.&Less classic and expensive, more funky and youthful attire is offered at virtually every doorstep along MelroseAvenue between La Brea and Fairfax.&Less classic and expensive, more funky and youthful attire is offered at virtually every doorstep along MelroseAvenue between La Brea and Fairfax.&On \textbf{tt-lanta-west} and mailwood in \textbf{La Brea-fen,La Brea-fen,La Brea-fen},the few doors leading to Fairfax\\ \hline
11.&They will lose over \$250,000.&They \textbf{gon lose} over \$250,000.&They will lose over \$250,000. \\ \hline
12.&I don't want them to follow in my footsteps, as I didn't go to college, but I want them to go.&I don't want them to follow in my footsteps, as \textbf{I ain't go to no college}, but I want them to go.&I don't want them to follow in my footsteps, as I didn't go to college, but I want them to go.\\ \hline
13.&The size is universal &Da \textbf{size universal}&The size is universal\\ \hline
14.&It coarsens and inflames dialogue in a way that tends to prevent exposure of which ideas are right and which are wrong.&It \textbf{coarsen} and \textbf{inflame} dialogue in \textbf{a way tend} to prevent exposure of which ideas right and which wrong.&Inadequate discussion of which ideas are correct and which are wrong is likely to prevent exposure to which ideas are correct\\ \hline
15.&Whatever they had done to patch him up hadn't been very successful, apparently.&Whatever they done to patch \textbf{his ass} up \textbf{ain't been} very successful, apparently.&not in a good way they seem to have it down his way\\
\bottomrule
\end{tabular}
}\caption{A comparison between sampled sentences from the original GLUE MNLI, and VALUE and STRAP transformed AAVE text. We see STRAP's creative phrase-level transformations (Examples 1-5), but also how STRAP can hallucinate and alter the meaning of the sentence (6-7), remove valuable details (8) and produce erratic behaviors (9-10). VALUE on the other hand appears to better capture AAVE morphosyntax (11-15).}
\label{tab:data_sample}
\end{table*}

\end{document}